\documentclass[letterpaper]{article}
\usepackage{aaai2027}
\usepackage[hyphens]{url}
\usepackage{graphicx}
\usepackage{natbib}
\usepackage{xcolor}
\usepackage{caption}
\usepackage{subcaption}
\usepackage{amsmath}
\usepackage{amsfonts}
\usepackage{booktabs}
\usepackage{array}
\usepackage{multirow}
\usepackage[most]{tcolorbox}
\usepackage{enumitem}
\usepackage{fvextra}
\graphicspath{{fig/}}

\copyrighttext{Under Review}

\title{From Social Coding to Agentic Coding: Productivity and Relational Reconfiguration in Open-Source Communities}
\author{%
  Mengying Zhou$^{1}$ \quad Yongjie Yin$^{2}$ \quad Yang Chen$^{2}$
}
\affiliations{%
  $^{1}$School of Computing and Artificial Intelligence, Shanghai University of Finance and Economics \\
  $^{2}$College of Computer Science and Artificial Intelligence, Fudan University
}

\begin{document}

\maketitle

\begin{abstract}
Open-source software communities are a form of digital public infrastructure that not only produces code, but also generates public knowledge and interpersonal relationships through visible collaboration. Generative coding agents (CAs) are an advanced tool to improve development efficiency while shifting part of activities from public human interaction to private human--agent loops. We study this shift using an LLM-based multi-agent simulation initialized with real GitHub data from 1,084 active developers and their repository relationships. After a warm-up with historical commits, we branch the same community state into parallel No-CA and CA conditions for 4-week simulations. CA introduction increases planned and completed tasks by 34.0\% and 39.0\%, respectively, and reduces median completion time from 45 to 20 minutes. However, adoption reaches only 26.0\%, and the gains concentrate among developers who are already more active and well connected. CAs also restructure task execution pathways. Direct human--human interaction declines from 32.4\% to 11.6\%, while CA-involved modes increase to 57.3\%, including 40.3\% completed through CA-assisted self-loops. Public knowledge generated under CA condition also provides less support for later tasks. On a standardized retrieval benchmark, the CA corpus achieves 22.3\% knowledge coverage, far below the 81.1\% achieved by the real-human corpus, and requires more retrieval steps with a lower success rate. These results reveal a productivity--public knowledge tension: coding agents increase technical production, but more work shifts to agent-mediated or private loops, leaving public records less useful to future contributors.
\end{abstract}

\section{Introduction}

Open-source software (OSS) has become a foundational form of digital public infrastructure. Operating systems, data science libraries, and scientific software are increasingly developed and maintained by distributed open-source communities~\cite{linaker2025advancing}.

OSS is not only a mode of code production but also a form of community organization that depends on sustained collaboration~\cite{dabbish2012social,mockus2002case}. Developers jointly complete technical tasks through pull requests, discussion, and code reviews~\cite{tsay2014influence}. Through these processes, they coordinate ongoing work, accumulate shared project knowledge, and recognize contributors’ expertise. Since these activities are usually publicly visible, the related information produced during tasks can be observed and reused by subsequent developers~\cite{jahanshahi2025beyond}. Therefore, the continued operation of OSS depends not only on code production but also on whether technical activities generate public knowledge and interpersonal connections.

Generative coding agents (CAs) are changing this process. Coding agents can assist with task understanding, code generation, debugging, testing, and revision, and have been shown to improve individual task-completion efficiency~\cite{weber2024significant}. However, unlike traditional OSS collaboration, interactions between developers and coding agents usually occur in private workspaces~\cite{mozannar2024reading}. The final commit or pull request may still be pushed to the public repository, while the intermediate revisions and process that led to it may remain private. Therefore, coding agents may change not only how efficiently technical work is completed but also how problems are solved and documented within the community~\cite{messeri2024artificial}.

Existing studies mainly examine the effects of CA-assisted programming on individual productivity, code quality, and developer experience, while paying less attention to community-level consequences~\cite{martinlopez2026more}. Particularly, it remains unclear whether access to direct technical assistance from coding agents shifts part of the activities from public human interaction to private human--agent collaboration, and whether this shift reduces the interpersonal exchange and public knowledge generated by each contribution~\cite{yonekawa2024analysis}. Considering OSS communities heavily rely on visible interaction to coordinate work, support newcomer learning, and build reusable knowledge, it is important to identify whether coding agents are beginning to reshape these underlying community mechanisms~\cite{hao2026artificial}.

Accordingly, this study examines how the introduction of coding agents changes technical production, interaction pathways, and public knowledge formation in OSS communities. We ask the following three research questions:

\begin{itemize}
    \item \textbf{RQ1:} How do community productivity and development efficiency change after coding agents are introduced, and how are adoption and productivity gains distributed across developers?
    \item \textbf{RQ2:} Do coding agents shift part of task-related activities from public human interaction to private human--agent collaboration, and what short-term relational changes accompany this shift?
    \item \textbf{RQ3:} How does coding-agent use affect the public knowledge generated through technical activity and its usefulness to subsequent developers?
\end{itemize}

To answer these questions, we construct an LLM-based multi-agent simulation initialized with real GitHub data, including 1,084 continuously active developers and their observed repository relationships. Each developer agent’s profile is constructed from real historical data, combining quantitative activity records with a qualitative biographical summary. The simulation consists of a 4-week warm-up and a 4-week intervention period. During the warm-up, real commits are injected chronologically so that agents learn developers' recent behavioral patterns through few-shot in-context learning. We then branch the same community snapshot into parallel No-CA and CA conditions, in which LLM-driven developer agents generate daily activities throughout the simulation. We conduct multiple independent runs for each condition to simulate community evolution.

The simulation results show that coding agents substantially increase technical production. Compared with No-CA, the number of planned tasks increases by 34.0\%, completed tasks increase by 39.0\%, and the median completion time decreases from approximately 45 to 20 minutes. However, these gains are unevenly distributed. Only 26.0\% of developers adopt a coding agent by the end of the simulation, and the productivity gains are concentrated among developers who are already more active and well-connected. 
Coding agents also restructure task-execution pathways. The proportion of tasks completed through direct human--human interaction decreases from 32.4\% under No-CA to 11.6\% under CA, while CA-involved modes account for 57.3\% of completed tasks. In particular, CA-assisted self-loops account for 40.3\%, indicating that a substantial amount of work moves into developer--agent loops.
A further concern lies in the retrievability of public knowledge. On a standardized query benchmark, the real-human corpus achieves 81.1\% knowledge coverage, whereas the size-matched CA corpus achieves only 22.3\%. Average retrieval steps increase from 2.63 to 8.02, while retrieval success decreases from 82.3\% to 22.3\%. Overall, coding agents increase technical production, but most activity is completed through agent-mediated private loops, and the resulting public records provide limited support for later tasks. In summary, our main contributions are as follows:

\begin{itemize}
    \item This work extends research on coding agents from individual development efficiency to the community level by examining the relationship between technical production and public collaboration. 
    \item This work develops a multi-agent simulation initialized with real community structures and distinguishes public human interaction from private human--agent collaboration. 
    \item This work shows that coding agents should be evaluated not only by productivity but also by how they reshape interaction pathways and public knowledge. In doing so, the study identifies early conditions through which agentic coding may affect the continued development of OSS communities.
\end{itemize}
\section{LLM-Based Simulation Setup and Intervention Design}
\label{sec:simulation_design}

Coding agents are primarily embedded in developers' local workflows. Therefore, their task delegation, coding assistance, and intermediate reasoning are not fully captured in public repositories~\cite{tufano2026developers}. Issues, Commits, and Pull Requests reflect final outputs but provide limited insight into how these outputs are produced. Moreover, observational data capture only one realized trajectory and cannot reveal how the same community would evolve under conditions with and without access to coding agents.

To address these limitations, we construct an LLM-based multi-agent simulation of an open-source software community grounded in real-world GitHub data~\cite{park2023generative,aher2023using,bonabeau2002agent}. Compared with traditional agent-based models that rely on hand-crafted behavioral rules~\cite{bonabeau2002agent}, LLM-based agents generate context-sensitive behaviors and natural-language interactions without exhaustive rule engineering, making them suited for modeling the open-ended, text-rich coordination characteristic of OSS development~\cite{xie2024trust}. 

\begin{figure*}[t]
    \centering
    \includegraphics[width=0.95\textwidth]{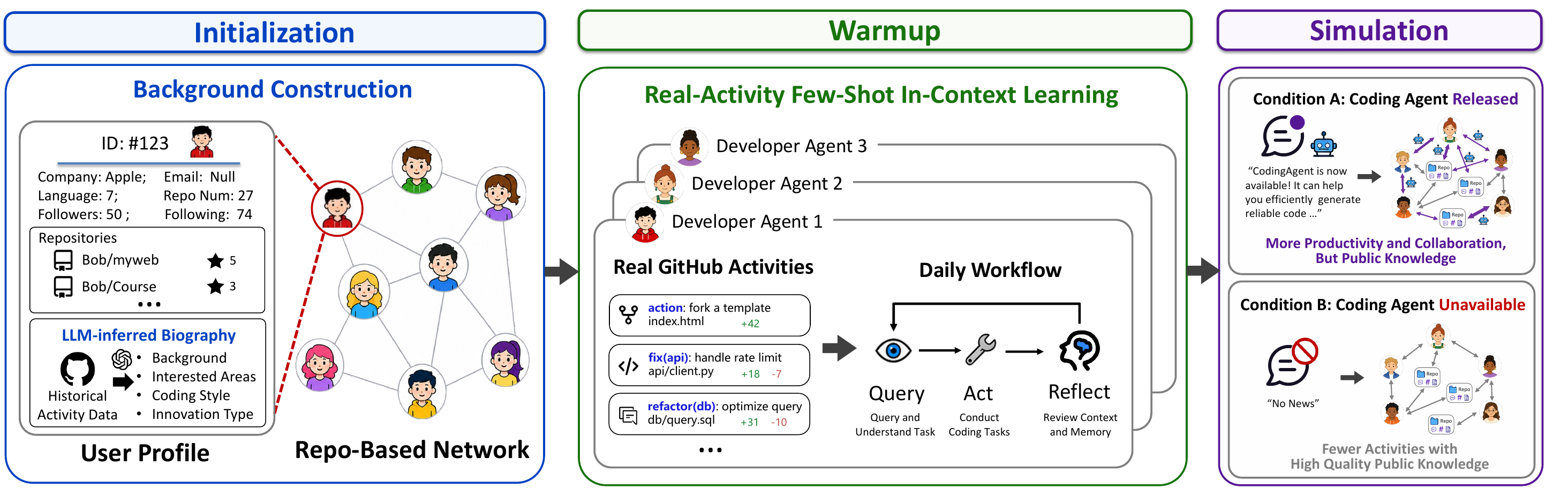}
    \caption{Overview of the Data-Grounded OSS Multi-Agent Simulation.}
    \label{fig:simulation_design}
\end{figure*}

As shown in Fig.~\ref{fig:simulation_design}, the overall workflow consists of three stages: \emph{Initialization}, \emph{Warmup}, and \emph{Simulation}. Initialization selects developers who are continuously active and have sufficient historical records, and then constructs the agent backgrounds and environment. The Warmup stage uses real-world activities as few-shot in-context examples to reconstruct developers’ recent behavioral states. Simulation then starts from the same community snapshot and compares the counterfactual outcomes when coding agents are unavailable and available.

\subsection{Initialization: Data Selection and Background Construction}

\paragraph{(1) Developer selection and collaboration graph construction.}
We first select developers who have sufficient data to support longitudinal and continuous behavioral modeling from the public GitHub Developer dataset~\cite{Gong2019MaliciousAccounts}. 

A developer must satisfy all of the following criteria: record at least one commit in each week from January 21 to March 18, 2018; have at least 50 historical commits; remain active from at least March 18, 2017 to March 18, 2018; own or contribute to at least three non-fork repositories; and have at least four non-empty profile fields. These criteria ensure sustained activity, sufficient behavioral history, meaningful project responsibility, and adequate information for constructing reliable LLM-based backgrounds. The final sample contains 1,084 developers. It covers diverse programming languages and development ecosystems and exhibits high long-term contribution intensity, broad cross-repository participation, and relatively stable recent activity.

Based on the selected developers, we construct a repository-based developer collaboration graph, in which an edge connects two distinct developers who have contributed to at least one common repository. The resulting network exhibits a realistic open-source collaboration structure, with one large connected component, several smaller components, and a number of isolated developers. This pattern reflects the uneven and decentralized nature of real-world OSS collaboration, where some developers are embedded in broader project networks while others remain active within smaller or independent contexts. We regard the sampled network as a reasonable representation of the underlying developer ecosystem. Detailed selection procedures, sample distributions, network statistics, and interactive visualizations are provided in the supplementary materials.

\paragraph{(2) Developer agent profile construction.}
Developer agent profile construction is an offline preprocessing procedure conducted outside the simulation timeline. Using all historical commit data collected before January 21, 2018, we construct an empirically grounded behavioral background for each developer. The background contains quantitative descriptions of contribution volume, activity periods, programming languages, repository participation, and contribution distributions, together with qualitative descriptions based on a biographical profile and the five-category classification of Innovation Diffusion Theory (IDT)~\cite{Rogers2003Diffusion}. The biographical profile is generated by an LLM from the developer's historical records and summarizes the developer's technical experience, common tasks, project responsibilities, and working habits. The IDT classification captures the developer's relatively stable tendency to adopt new technologies~\cite{Rogers2003Diffusion}.

\subsection{Warmup: Few-shot In-Context Learning with Real Activities}
\paragraph{(1) Purpose of real-activity injection.}
The Warmup stage uses real-activity few-shot in-context learning (ICL). We inject four weeks of real commit activity (January 21–February 18, 2018) into the corresponding agents’ contexts as behavioral examples to reconstruct developers' recent project states and work patterns before the intervention simulation.

The profiles constructed during Initialization capture developers' long-term and relatively stable technical backgrounds, whereas real-activity ICL provides recent project contexts and behavioral patterns. Together, they portray each agent's behavior as consistent with both long-term characteristics and recent work patterns.

\paragraph{(2) Design of developer agents and daily workflow.}
The simulation is implemented using AgentSociety~\cite{Piao2025AgentSociety}. Each developer is modeled as an LLM agent with a background profile, repository relationships, current tasks, and dynamic memory. As shown in Fig.~\ref{fig:simulation_design}, each simulation day follows three steps:

\begin{enumerate}
    \item \textbf{Query:} Inspect related repository states, finished tasks, project responsibilities, and planned work.
    \item \textbf{Act:} Select and perform development activities, including creating tasks, maintaining repositories, and discussing issues. In the CA branch, the coding agent can perform any of these actions on the developer’s behalf.
    \item \textbf{Reflect:} Integrate the day’s context and outcomes into memory and update action preferences.
\end{enumerate}

During Warmup, each developer agent receives the number and message content of its real commit records and executes the corresponding activities. It then incorporates the updated task states and recent information during Reflect. Through continuous real-activity injection, the agent develops recent project context, activity patterns, and memory.

\subsection{Simulation: Counterfactual Simulation and Coding Agent Intervention}

During the simulation, developers continue to follow the same three-step workflow used in Warmup. Starting from the same post-Warmup snapshot, we conduct two parallel simulation branches: No-CA and CA conditions. The only difference between the two conditions is whether developers can use a coding agent as an execution channel during Act. 

\paragraph{(1) No-CA condition.}
In the No-CA condition, the system prompt explicitly states that coding agents are unavailable, preventing the underlying LLM from implicitly assuming their existence based on pretraining knowledge. Developer agents maintain their own repositories or public repositories from others, with all actions performed directly by themselves.

\paragraph{(2) CA condition.}
At the start of the CA condition, the platform announces the availability of the coding agent, emphasizing its ability to help produce reliable code and improve development efficiency. Based on the assumption that developers with broader exposure to technical ecosystems and higher recent activity are more likely to adopt new technologies early, we define developers who have used at least seven programming languages and completed at least 15 commits during Warmup as seed adopters. 
Once seed developers use the coding agent, their CA-assisted activities become visible to others through the collaborated repositories. Subsequent adoption depends on each developer's IDT category, planned tasks, and prior experience.

Each simulation spans four weeks, corresponding to the real-world period from February 19 to March 18, 2018. To reduce variation from LLM sampling and stochastic multi-agent interactions, we independently run each condition three times under the same configuration and report the aggregated results. This design balances computational cost and result stability, consistent with prior LLM-based simulation studies~\cite{park2023generative,aher2023using}. Our primary experiments use DeepSeek V4, and the main conclusions remain consistent when agents are powered by GLM-5.1 or Qwen3. A broader cross-model comparison covering additional model families, including GPT and Gemini, is left for future work due to computational cost constraints. All simulation prompts are provided in the supplementary materials for reproducibility.
\section{Results}

We organize the results around four questions: whether the simulation reproduces baseline community activity, how coding agents affect productivity and its distribution, how they reshape direct cross-developer interaction, and how they influence the production and task relevance of public knowledge. Together, these analyses reveal a tension between increased productivity and reduced public collaboration and knowledge sharing, as more activity shifts to private, agent-mediated interactions.

\subsection{Simulation Validity}

Before conducting the result analysis, we evaluate whether the simulation reproduces the overall activity level and developer heterogeneity observed in the real community. This validation does not require the simulation to reproduce individual events, but examines whether it preserves the overall activity patterns and  interaction structures.

First, we compare simulated and empirical activity at the developer--day level with MAE and RMSE metrics, respectively. The results show that the MAE and RMSE are $1.86$ and $3.80$. Moreover, the mean total activities per developer is $31.0$ in the empirical data and $32.2$ in the simulation, while the corresponding medians are $28$ and $25$. These results show that the simulation closely preserves the overall activity patterns.

We further validate developer heterogeneity across three metrics: 1) activity Gini measures inequality in total activities across developers; 2) active-repository Gini captures differences in the number of repositories to which developers contribute, and 3) repository-concentration Gini measures whether contributions are concentrated in a small number of repositories. The $\Delta$Gini values for these metrics are 0.1138$\pm$0.0101, 0.1085$\pm$0.0081, and 0.0802$\pm$0.0166, respectively, with a mean $\Delta$Gini of 0.1009$\pm$0.0078. This indicates that the simulation broadly preserves the multidimensional heterogeneity of the real community, although it slightly amplifies the concentration of activity and project participation. Overall, the simulation closely reproduces the community's activity patterns and developer heterogeneity, providing a reasonable controlled environment for the subsequent counterfactual analysis.

\begin{figure}[t]
    \centering
    \begin{subfigure}[b]{0.75\linewidth}
        \centering
        \includegraphics[width=\linewidth]{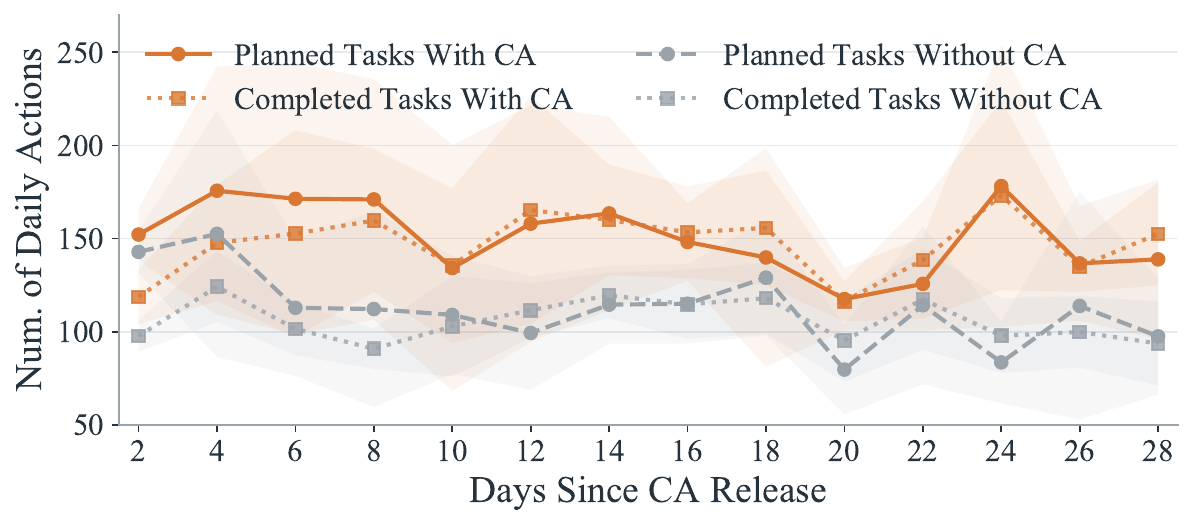}
        \caption{Planned and Completed Tasks}
        \label{fig:task_output}
    \end{subfigure}
    \hfill
    \begin{subfigure}[b]{0.24\linewidth}
        \centering
        \includegraphics[width=\linewidth]{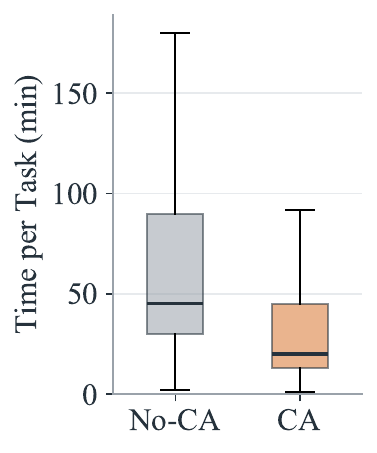}
        \caption{Task Completion Time}
        \label{fig:time_cost}
    \end{subfigure}
    \caption{Task Productivity and Time Efficiency Under No-CA and CA Conditions.}
    \label{fig:productivity}
\end{figure}

\subsection{RQ1: Productivity Improvement and Unequal CA Adoption}

\paragraph{(1) Higher productivity and efficiency.}
The introduction of coding agents substantially increases productivity in the simulated community. As shown in Figure~\ref{fig:task_output}, both planned and completed tasks are generally higher under the CA condition than under the No-CA condition. The cumulative number of planned tasks increases from 3,151 under No-CA to 4,221 under CA, while the number of completed tasks increases from 2,969 to 4,128, corresponding to increases of 1.34x and 1.39x, respectively. Meanwhile, Figure~\ref{fig:time_cost} shows that the median task completion time decreases from approximately 45 to 20 minutes. Particularly, the 75th-percentile completion time for longer tasks also decreases from 90 to 45 minutes. These results suggest that coding agents not only encourage developers to take on more tasks but also improve execution efficiency, consistent with the phenomenon observed in real-world development settings~\cite{ziegler2024measuring,weber2024significant}.

\begin{figure}[t]
    \centering
    \begin{subfigure}[t]{1\linewidth}
        \centering
        \includegraphics[width=\linewidth]
        {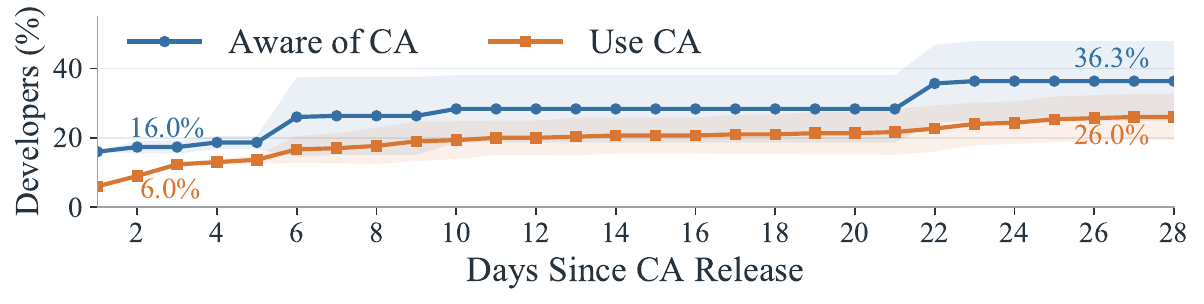}
        \caption{CA Awareness and Adoption}
        \label{fig:ca_awareness_adoption}
    \end{subfigure}
    \begin{subfigure}[t]{1\linewidth}
        \centering
        \includegraphics[width=\linewidth]
        {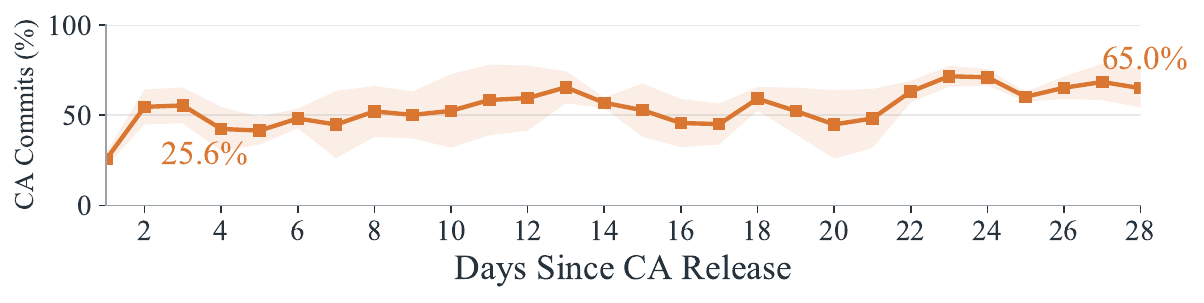}
        \caption{CA-Assisted Commits Ratio}
        \label{fig:ca_ratio}
    \end{subfigure}
    \begin{subfigure}[t]{1\linewidth}
        \centering
        \includegraphics[width=\linewidth]
        {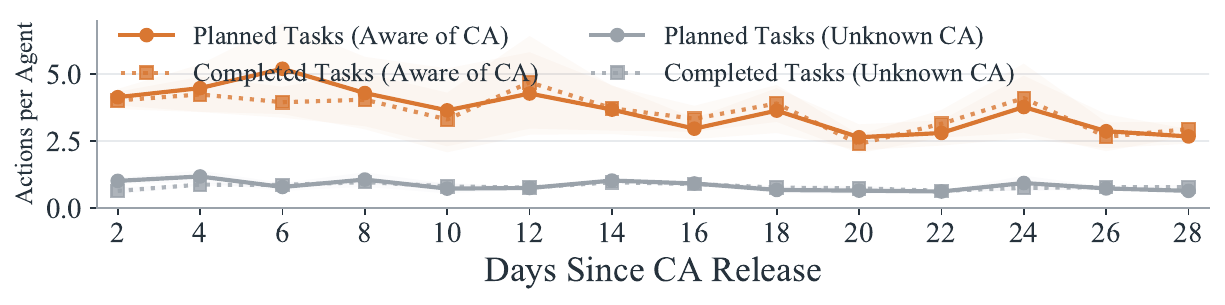}
        \caption{Tasks per Developer by Awareness}
        \label{fig:per_capita_awareness}
    \end{subfigure}
    \caption{Unequal Diffusion and Intensification of CA Adoption.}
    \label{fig:ca_diffusion}
\end{figure}

\paragraph{(2) CA diffusion and use concentrate among already active developers.}
However, awareness and adoption of CAs do not spread evenly across the community. 
As shown in Figure~\ref{fig:ca_awareness_adoption}, CA awareness increases from 16.0\% to 36.3\%, while adoption rises from 6.7\% to 26.0\%. 74.0\% of developers still do not use CAs by the final simulation day, indicating that CA diffusion remains limited to a minority of the community. 
Despite this limited adoption, CA-assisted commits account for an increasingly disproportionate amount of community production, rising from 25.6\% to 65.0\% of all commits, as shown in Figure~\ref{fig:ca_ratio}. To further examine this phenomenon, Figure~\ref{fig:per_capita_awareness} depicts that developers who are aware of CAs consistently plan and complete markedly more tasks per capita than those who remain unaware. Together, these results reveal a clear ``participation amplifier'' effect under the current active-developer-led diffusion setting. A relatively small group of CA-adopting developers uses CAs intensively and contributes a disproportionate contribution of development activity, indicating that CA-related productivity gains are concentrated among developers who are already more active and well connected.

\subsection{RQ2: Agent Mediation and the Thinning of Direct Cross-Developer Interaction}

\begin{figure}[t]
    \centering
    \includegraphics[width=1\linewidth]{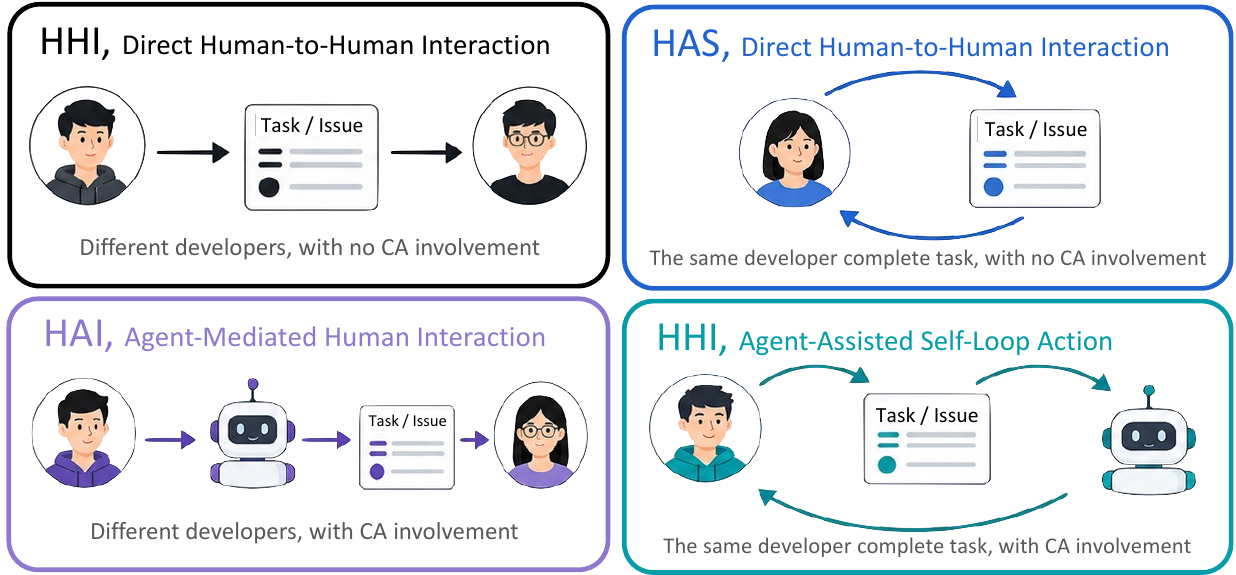}
    \caption{Four Task-Level Interaction Modes}
    \label{fig:interaction_illustration}
\end{figure}

\paragraph{(1) Task execution shifts from direct human interaction toward agent mediation and self-contained loops.}
To answer RQ2, we classify each completed task into one of four activity modes, HHI, HSA, AHI, and ASA, according to whether the task crosses developer boundaries and whether a coding agent participates. Their definitions and illustrations are provided in Figure~\ref{fig:interaction_illustration}. Based on this classification, we examine the fraction of completed tasks represented by each interaction mode.

\begin{figure}[t]
    \centering
    \includegraphics[width=\linewidth]{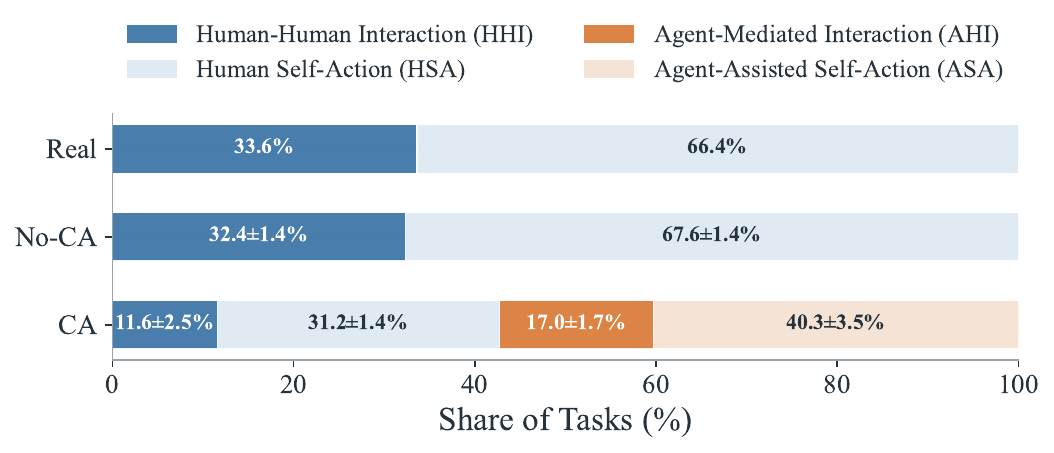}
    \caption{Task Composition by Interaction Mode.}
    \label{fig:task_composition}
\end{figure}

Figure~\ref{fig:task_composition} shows the composition of task modes across conditions. In both the real data and the No-CA condition, tasks consist only of HHI and HSA, and their compositions are broadly similar. Under the No-CA condition, HHI and HSA account for 32.4\%$\pm$1.4\% and 67.6\%$\pm$1.4\% of completed tasks, respectively. The task composition changes substantially under the CA condition. The proportion of HHI decreases to 11.6\%$\pm$2.5\%, while HSA decreases to 31.2\%$\pm$1.4\%. At the same time, AHI and ASA account for 17.0\%$\pm$1.7\% and 40.3\%$\pm$3.5\%, respectively. In total, 57.3\% of completed tasks involve a CA, with ASA becoming the largest single task mode. This pattern shows that coding agents not only participate in cross-developer tasks but also allow a large amount of tasks to be completed within developer--agent loops.

Notably, the overall proportion of cross-developer tasks decreases moderately, from 32.4\% HHI under No-CA to 28.6\% combined HHI and AHI under CA, while its internal composition changes markedly. Among these 28.6\% of cross-developer tasks under the CA condition, 59.9\%$\pm$4.4\% of tasks involve a CA. In other words, CAs do not cause cross-developer tasks to disappear at scale. Instead, they shift a substantial amount of these tasks from direct HHI to agent-mediated AHI. Therefore, the most insightful finding observed in RQ2 is the reduction of direct human participation and the increasing mediation of cross-developer tasks, rather than a simple decline in the total amount of cross-developer interaction.

\paragraph{(2) Relational breadth and short-term continuity remain broadly stable.} Beyond task modes, we examine interaction diversity and short-term continuity at the developer-pair level. \texttt{Context Breadth} measures the average number of distinct repository--interaction-type contexts associated with each interacting developer pair. A higher value indicates that a pair interacts through more activity types or across more repositories. \texttt{Repeated Interaction Rate} measures the proportion of interacting pairs that reconnect across at least two tasks and two different days. Because the 4-week observation window is short, the latter captures short-term interaction continuity rather than long-term relationship persistence.

Table~\ref{tab:interaction_metrics} summarizes the interaction indices at the developer-pair level. Context breadth is 6.60$\pm$0.31 under No-CA and 7.05$\pm$0.76 under CA, indicating that interactions may even broaden slightly across repositories and interaction types after CA introduction. This is also reflected in the increase in developer pairs that interact repeatedly across tasks and days, from 23.0$\pm$2.2 to 24.0$\pm$0.1. Repeated interaction rate decreases only slightly, from 93.3\%$\pm$3.7\% to 91.2\%$\pm$1.6\%. 
These results suggest that coding agents primarily reshape how cross-developer tasks are carried out, without immediately weakening the breadth or short-term continuity of developer relationships.

\begin{table}[!t]
    \centering
    \small
    \caption{Developer-Pair Interaction Breadth and Short-Term Continuity}
    \label{tab:interaction_metrics}
    \begin{tabular}{lcc}
        \toprule
        \textbf{Metric} & \textbf{No-CA} & \textbf{CA} \\
        \midrule
        Context Breadth & 6.60$\pm$0.31 & 7.05$\pm$0.76 \\
        Repeated Pairs  & 23.0$\pm$2.2    & 24.0$\pm$0.1   \\ 
        Repeated Interaction Rate (\%) & 93.3$\pm$3.7 & 91.2$\pm$1.6 \\
        \bottomrule
    \end{tabular}
\end{table}

\subsection{RQ3: Public Knowledge Retrieval}

To answer RQ3, we examine whether public knowledge records produced under the CA condition can support subsequent development tasks. These public knowledge includes text associated with public activities such as issues and commits, but exclude content generated only within private developer--agent ASA loops. We evaluate their semantic coverage and retrieval efficiency for future project tasks.


\paragraph{(1) Public records produced under the CA condition provide limited semantic coverage.} 
To assess whether public knowledge remains useful to later participants, we evaluate whether corpora constructed from public information under different conditions can support a standardized set of tasks. This setting represents a newcomer who relies only on existing public records to understand project-specific work and cannot use a CA or seek interactive help from existing developers. The benchmark consists of 8,822 real commits produced between March 19 and May 19, 2018, none of which are observed or used by the agents.

We evaluate two corpus conditions. The empirical baseline consists of 2,422 real human commits produced between February 19 and March 18, 2018. For a task set $\mathcal{Q}$ and a public knowledge corpus $\mathcal{C}$, we define \texttt{Public Knowledge Coverage} as the proportion of query tasks for which a semantically matching record can be found in the corpus. For each query task, we calculate the TF--IDF cosine similarity between the query and every corpus record. A query is considered covered when its similarity exceeds 0.3. Under the real human corpus, public knowledge coverage reaches 81.1\%. 
To keep a fair comparison, we sample the same number of records from the CA-stage corpus as in the real human corpus and repeat this process five times. We find that the coverage is only 22.3\%$\pm$2.2\% under the CA condition, representing a relative decrease of 72.5\% from the real-data baseline.

\paragraph{(2) Locating relevant knowledge requires more retrieval steps in the CA corpus.} 
We further introduce \texttt{Average Retrieval Steps} and \texttt{Retrieval Success Rate} to measure the efficiency of a newcomer-like retrieval process. For each query task, the system uses a repository-level reference text as the initial query and returns the Top-5 unseen records in each round. If the best result has a TF--IDF cosine similarity above 0.3, the round is counted as a successful hit. Otherwise, the eight most distinctive TF--IDF terms from the best-matching record are added to the query, and retrieval continues for up to ten rounds. Average retrieval steps is the mean number of rounds across 300 sampled queries, while retrieval success rate is the proportion of queries that achieve a successful hit within ten rounds.

In the real human corpus, average retrieval steps is 2.63 and retrieval success rate is 82.3\%. In the size-matched CA corpus, average retrieval steps increases to 8.02$\pm$0.10, a rise of 205.1\%, while retrieval success rate falls to 22.3\%$\pm$1.4\%. Together, these results show that although the CA-stage corpus is larger than the real corpus, the additional records do not provide broader knowledge coverage. Their semantic coverage and retrievability remain limited, requiring more rounds of query expansion, while most queries still fail to find a match above the threshold.

\section{Discussion}

OSS communities produce not only software but also public records of how problems are solved~\cite{benkler2002coases,lakhani2003open}. Discussions, code reviews, revision histories, and technical explanations allow knowledge to be reused beyond the original task~\cite{dabbish2012social,feng2026charting}. Coding agents alter this process because assistance often occurs privately. This creates a gap between access to knowledge and contribution to public knowledge. Developers may receive fast, personalized support, but that help does not automatically become useful to the wider community~\cite{messeri2024artificial}.  Similar tensions arise in open science, Wikipedia, and public question-answering communities~\cite{hao2026artificial,messeri2024artificial} as well. 

This work also suggests that OSS health should be assessed beyond visible productivity. OSS communities facilitate social learning through interpersonal interaction, particularly for newcomers to learn project norms, receive feedback, gain recognition, and take on greater responsibility. If coding agents replace too much of this early interaction, they may improve short-term productivity while weakening mentoring and social learning. Similar concerns are increasingly discussed in education, where AI assistance may boost immediate performance while weakening active problem solving, feedback seeking, and peer learning~\cite{bastani2025generative}.

Generative agents can strengthen individual capabilities while reducing the visible exchanges through which communities share norms, recognize expertise, and support newcomer onboarding. The broader issue is therefore not simply whether AI replaces people, but where coordination and learning occur, who can observe them, and who controls the resulting records~\cite{feng2026charting,benkler2002coases}. Although these implications remain hypotheses beyond the simulated OSS setting, they suggest that evaluations of AI-assisted work should consider public knowledge and participation alongside individual productivity.

\section{Related Work}

\noindent\textbf{AI-Assisted Programming and Agentic Coding. }
Research on AI-assisted programming has progressed from neural code-generation models to LLM-based coding agents capable of autonomous task execution. Foundational neural architectures established techniques for translating natural language to source code~\cite{sun2020treegen,wang2021code}. Studies of large language model-based tools such as GitHub Copilot report substantial improvements in task completion, development speed, and perceived productivity, although the benefits vary across developers and task types~\cite{ziegler2024measuring,weber2024significant,zhu2024hot}. Interaction studies further show that developers use AI assistants both to accelerate familiar work and to explore unfamiliar code, while still needing to inspect, validate, and revise generated outputs~\cite{barke2023grounded,mozannar2024reading}.

Recent work suggests that greater automation reshapes not only development efficiency but also the organization of developer work. AI assistants can reduce manual coding but may also weaken code understanding, code ownership, and knowledge transfer~\cite{welter2025knowledge,martinlopez2026more}. Emerging studies have examined self-reported AI use in OSS, AI-assisted pull requests, downstream maintainability, and the concentration of development activity within private AI-mediated workflows~\cite{tufano2026developers,borg2026echoes,yonekawa2024analysis}. A recent socio-technical framework further maps how GenAI reshapes OSS communities across development practices, documentation, community engagement, and governance, highlighting tensions between productivity gains and community sustainability~\cite{feng2026charting}. However, most existing work remains centered on individual performance or project-level artifacts, with limited examination of how agentic coding reshapes community-wide interaction structures and public knowledge dynamics.

\noindent\textbf{Open Source as Social Coding and Relational Knowledge Infrastructure. }
OSS development has long been understood as a form of social coding in which technical production is organized through visible human interaction. GitHub's transparency allows developers to observe contributions, infer expertise, coordinate work, and evaluate potential collaborators~\cite{dabbish2012social}. Pull-request evaluation likewise depends not only on technical quality, but also on prior participation, discussion, and the contributor's relationship with project members~\cite{tsay2014influence}. Human interaction therefore forms part of the relational infrastructure of OSS, increasingly recognized as digital public infrastructure underpinning modern software supply chains~\cite{linaker2025advancing}.

This visibility also supports community continuity. Newcomers often begin with bounded tasks, learn project practices through observation and feedback, and gradually move toward more central roles~\cite{vonkrogh2003community}. At the same time, software collaboration produces knowledge as well as code. Code review, issue discussion, and revision processes communicate design rationales, project conventions, and implementation knowledge across contributors~\cite{bacchelli2013expectations,caulo2020knowledge}. Because these interactions remain publicly visible, their value can extend beyond the original participants and be reused by future contributors.

This view is consistent with research on communities of practice and online knowledge collaboration, which treats learning as embedded in social participation and knowledge as jointly produced through interaction~\cite{brown1991organizational,wenger2000communities,wasko2005share,faraj2011knowledge}. Coding agents challenge this model by moving explanation, debugging, and implementation guidance into private human--AI conversations, potentially reducing the public visibility that underpins OSS knowledge reuse and community reproduction~\cite{feng2026charting}. Such assistance may improve individual access to knowledge, but it does not automatically become publicly reusable community memory. Recent studies have begun to document the growing use of conversational AI as a source of technical knowledge~\cite{welter2025knowledge}, yet its effects on community-level knowledge accumulation and relational reproduction remain largely unexplored.

\section{Conclusion}

By implementing a multi-agent simulation powered by LLMs and initialized with real GitHub data, this study reveals that coding agents increase OSS community productivity while reshaping task-execution pathways and public knowledge production and quality. The simulation results show that CAs can increase task generation, completion, and speed. However, adoption remains limited, and productivity gains concentrate among developers who are already active and well connected. Meanwhile, direct human-human interaction declines as more work shifts to agent-mediated and self-contained loops. Moreover, public knowledge generated under the CA condition provides lower knowledge coverage and retrieval success than a real human corpus. These findings suggest that coding agents should be evaluated not only by productivity and efficiency, but also by how they redistribute participation, reorganize collaboration, and affect the accumulation of public knowledge.

This study is limited by its fixed developer population and 8-week observation window, which restrict the analysis of longer-term community evolution. Future work should therefore extend the simulation to larger and more diverse communities, allow developers to enter and leave endogenously, model transitions between roles, and examine longer time horizons. These extensions would enable a more direct investigation of community dynamics.

\bibliography{ref}

\section{Supplementary Material}

\subsection{1. Code and Data Availability}

The code and corresponding data  will be released later.


\begin{table*}[!t]
    \centering
    \small
    \caption{The core set of indicators corresponding to the three research questions.}
    \label{tab:cross_model_core}
    \begin{tabular}{l|l|ccc}
        \hline
        \textbf{RQ} & \textbf{Metric} & \textbf{DeepSeek-V4} & \textbf{GLM-5.2} & \textbf{Qwen3.7} \\
        \hline
        \multirow{3}{*}{RQ1}
        & Completed Tasks change (\%) & +39.0 & +86.1 & +17.3 \\
        & Median Task Time change (\%) & $-55.6$ & $-41.7$ & $-40.0$ \\
        & Final CA Adoption (\%) & $26.0 \pm 6.7$ & 28.0 & 28.0 \\
        \hline
        \multirow{2}{*}{RQ2}
        & HHI share under CA (\%) & 12.8 & 9.1 & 11.9 \\
        & Agent Mediation Rate under CA (\%) & 56.0 & 72.2 & 67.0 \\
        \hline
        \multirow{2}{*}{RQ3}
        & Knowledge Coverage (\%; Real/CA, $\Delta$pp) & $67.6/13.6\;(-53.9)$ & $70.5/8.0\;(-62.5)$ & $67.9/6.3\;(-61.6)$ \\
        & Avg. Retrieval Steps (Real/CA, $\Delta$) & $3.54/8.75\;(+5.21)$ & $3.21/9.34\;(+6.13)$ & $3.54/9.45\;(+5.91)$ \\
        \hline
    \end{tabular}
\end{table*}

\subsection{2. Cross-Model Consistency and Robustness Checks}
\label{sec:supp_results}

This section examines whether the main findings depend on the LLM used to drive developer agents. We focus on whether the direction of the observed effects remains consistent across models, while allowing their magnitude to vary. The purpose is therefore to assess the robustness of the conclusions rather than to compare model performance.

\subsubsection{2.1 LLM Configuration and Evaluation Design}
\label{sec:supp_repro}
\label{sec:supp_cross_model_design}

The main experiments use DeepSeek-V4, with three independent runs for each condition. We additionally conduct robustness checks using GLM-5.2 and Qwen3.7. For each alternative model, we run one paired No-CA/CA experiment using the same post-Warmup community state, developer population, platform mechanisms, timeline, and parameter settings. The three models were released in close temporal proximity: DeepSeek-V4 on April 24, 2026; Qwen3.7 on May 19, 2026; and GLM-5.2 on June 16, 2026. All were publicly accessible at the time the experiments were conducted.

To account for differences in the baseline behavior of each LLM, we compare the CA and No-CA conditions within each model. We retain a small set of indicators corresponding to the three research questions: completed tasks, completion time, and CA adoption for RQ1. HHI share and Agent Mediation Rate for RQ2. And Public Knowledge Coverage and Average Retrieval Steps for RQ3.

\subsubsection{2.2 Consistency of Core Findings Across Models}
\label{sec:supp_cross_model}

Table~\ref{tab:cross_model_core} reports the core set of indicators corresponding to the three research questions. Across models, the results consistently show higher productivity and shorter completion time, limited CA adoption, increased agent mediation, and a substantial public-knowledge retrieval gap between CA-generated and real-human records.

\paragraph{(1) Production, Efficiency, and Adoption.}
All three models show higher technical production and shorter completion time after CA introduction. Completed Tasks increase by 39.0\% under DeepSeek-V4, 86.1\% under GLM-5.2, and 17.3\% under Qwen3.7. The substantially larger increase under GLM-5.2 likely reflects its stronger coding-oriented training and response tendency. GLM-5.2 is designed for coding, agentic engineering, and long-horizon task execution, which may make its agents more likely to decompose repository work into actionable tasks and make fuller use of CA assistance. As a result, CA availability produces a larger increase in both task planning and completion under GLM-5.2. By contrast, DeepSeek-V4 and Qwen3.7 appear more conservative in expanding the amount of planned work. These differences affect the magnitude of the estimated productivity gain but not its direction: all three models indicate that CA introduction increases completed technical work. Because the alternative models are evaluated through single paired runs, the particularly large GLM-5.2 estimate should be interpreted as a model-specific response pattern rather than a precise estimate of the general CA effect.

\paragraph{(2) Task-Execution Pathway.} The restructuring of task execution is also consistent across models. Under the CA condition, HHI accounts for only 12.8\% of completed tasks under DeepSeek-V4, 9.1\% under GLM-5.2, and 11.9\% under Qwen3.7. Among cross-developer tasks, the Agent Mediation Rate reaches 56.0\%, 72.2\%, and 67.0\%, respectively. Thus, direct human--human execution remains a small component of completed work, while more than half of cross-developer tasks involve CA mediation under every model. Although the degree of mediation varies, the models consistently reproduce the shift from direct human interaction toward agent-mediated task execution.

\paragraph{(3) Public Knowledge Retrieval.} The gap between CA-generated and real-human public knowledge persists across all three models. Knowledge Coverage decreases from 67.6\% to 13.6\% under DeepSeek-V4, from 70.5\% to 8.0\% under GLM-5.2, and from 67.9\% to 6.3\% under Qwen3.7. At the same time, Average Retrieval Steps increase by 5.21, 6.13, and 5.91 steps, respectively. These consistent differences show that the lower coverage and greater retrieval effort observed for CA-generated records are not specific to a single LLM backend.

\subsubsection{2.3 Summary}
\label{sec:supp_cross_model_scope}

The cross-model checks support the consistency of the main conclusions. Across all three LLMs, CA introduction increases completed work and reduces completion time, adoption remains limited, and cross-developer task execution becomes increasingly agent-mediated. Moreover, CA-generated public records provide substantially lower knowledge coverage and require greater retrieval effort than real-human records.

The magnitude of these effects remains model-dependent. GLM-5.2 produces the largest increase in completed tasks and the highest Agent Mediation Rate, while Qwen3.7 exhibits the largest public-knowledge coverage gap. However, none of these differences reverses the central findings.
The robustness analysis remains limited by the cost. A broader evaluation covering additional model families, including GPT and Gemini, is left for future work because of the substantial computational cost.

\begin{table*}[htbp]
    \centering
    \footnotesize
    \caption{Developer-selection criteria.}
    \label{tab:selection_criteria}
    \begin{tabular}{p{0.04\linewidth}|p{0.25\linewidth}|p{0.33\linewidth}|p{0.32\linewidth}}
        \hline
        No. & Criterion & Operational definition & Methodological purpose \\
        \hline
        1 & Continuous recent activity & At least one commit in every week from April 2 to May 27, 2018 & Identifies developers who remain active throughout the eligibility period. \\\hline
        2 & Sufficient historical activity & At least 50 commits in the screening record & Provides enough observations to estimate behavioral patterns. \\\hline
        3 & At least one year of observable activity & Activity begins no later than May 27, 2017, and continues through at least May 27, 2018 & Excludes temporary and one-time contributors. \\\hline
        4 & Cross-project participation & Participation in at least three non-fork repositories & Ensures meaningful exposure to multiple projects and repository roles. \\\hline
        5 & Sufficient public profile information & At least four non-empty profile fields & Supports the construction of developer profiles based on real data. \\
        \hline
    \end{tabular}
\end{table*}
\subsection{3. Data Selection and Descriptive Statistics}
\label{sec:supp_sample}

This section describes the data source and observation windows, explains the developer-selection procedure, and summarizes the technical composition, activity patterns, repository participation, and collaboration structure of the final sample. These analyses establish whether the selected developers provide sufficiently rich and heterogeneous histories for agent initialization while preserving the network variation required to examine changes in production, interpersonal collaboration, and public knowledge exchange. The final sample contains 1,084 persistent developers who produce 369,645 commits during the descriptive Analysis window, with a mean of 341.0 commits per developer and a median of 322. The sample is intentionally activity-focused and represents a typical group of GitHub developers.

\subsubsection{3.1 Data Source and Sample Selection}
\label{sec:supp_data_source}

\paragraph{(1) Raw Data Source.}

Developer selection is based on the public GitHub Developer Dataset~\cite{Gong2019MaliciousAccounts}, which provides commit histories, repository-participation records, and public profile information for approximately 329,000 developers.

\paragraph{(2) Observation Windows.}

All records are organized into three consecutive windows with distinct methodological purposes:

\begin{itemize}
    \item \textbf{Historical Background Period} (October 1, 2014--January 21, 2018): historical commits, repository participation, and profile information used to construct developer backgrounds, estimate long-term activity patterns, and build the initial collaboration network.
    \item \textbf{Warmup Period for In-Context Learning (ICL)} (January 22--February 18, 2018. Four weeks): recent real activities replayed as few-shot behavioral demonstrations to update agent memory, activity statistics, and platform states before the experiment.
    \item \textbf{Simulation Period} (February 19--March 18, 2018. Four weeks): the counterfactual experimental period in which agents operate autonomously under the No-CA and CA conditions.
\end{itemize}

The Historical Background Period provides the real data basis for developer profiling and community initialization, the Warmup ICL Period bases agents on recent observed behavior, and the Simulation Period is used to generate and compare experimental outcomes.

\paragraph{(3) Developer-Selection Criteria.}

We select developers with sufficiently rich, continuous, and cross-project activity records to support longitudinal agent modeling. Each developer must satisfy all five criteria in Table~\ref{tab:selection_criteria}.

The criteria are applied sequentially to the source data, producing a final analytical sample of 1,084 developers.

\subsubsection{3.2 Descriptive Statistics}
\label{sec:supp_descriptive}

\paragraph{(1) Sample Composition.}
\label{sec:supp_composition}

The selected 1,084 developers span a broad range of software ecosystems (Table~\ref{tab:language_distribution}). JavaScript and TypeScript form the largest language group, but no language group represents a majority of the sample. Developers with maintainer or mixed contributor--maintainer roles account for most of the cohort (Table~\ref{tab:role_distribution}), preserving variation in both technical background and project responsibility.

\begin{table}[htbp]
    \centering
    \small
    \caption{Distribution of developers by primary programming-language group.}
    \label{tab:language_distribution}
    \begin{tabular}{l|r|r}
        \hline
        Language group & Developers & Percentage \\
        \hline
        JavaScript/TypeScript & 347 & 32.0\% \\
        Python & 171 & 15.8\% \\
        Java & 141 & 13.0\% \\
        C/C++/C\# & 108 & 10.0\% \\
        Other & 102 & 9.4\% \\
        PHP/Ruby & 84 & 7.7\% \\
        Go/Rust/Swift/Objective-C & 81 & 7.5\% \\
        Web (HTML/CSS) & 50 & 4.6\% \\
        \hline
        Total & 1,084 & 100.0\% \\
        \hline
    \end{tabular}
\end{table}

\begin{table}[htbp]
    \centering
    \small
    \caption{Distribution of developers by repository role.}
    \label{tab:role_distribution}
    \begin{tabular}{l|r|r}
        \hline
        Developer role & Developers & Percentage \\
        \hline
        Contributor & 268 & 24.7\% \\
        Maintainer & 367 & 33.9\% \\
        Mixed contributor--maintainer & 449 & 41.4\% \\
        \hline
        Total & 1,084 & 100.0\% \\
        \hline
    \end{tabular}
\end{table}

\paragraph{(2) Activity and Commit Distributions.}
\label{sec:supp_activity}

For descriptive purposes, the Historical Background Period, Warmup ICL Period, and Simulation Period together form the Full Analysis Period (October 1, 2014--March 18, 2018). The statistics in the following subsection summarize the developer activity across this complete period. Real activities observed during the Simulation Period are used only for sample description and are not provided to agents during the  simulation.

The sample consists primarily of persistent and productive developers but retains substantial variation in contribution volume and temporal regularity. Table~\ref{tab:commit_distribution} summarizes the distribution of commits across the Full Analysis Period.

\begin{table}[htbp]
    \centering
    \small
    \caption{Distribution of commits per developer during the Full Analysis Period (October 1, 2014--March 18, 2018).}
    \label{tab:commit_distribution}
    \begin{tabular}{l|r}
        \hline
        Statistic & Total commits \\
        \hline
        Minimum & 1 \\
        25th percentile & 178 \\
        Median & 322 \\
        75th percentile & 494 \\
        Maximum & 6,521 \\
        Mean & 341.0 \\
        \hline
        Developers with $\geq 50$ commits & 100\% \\
        Developers with $\geq 100$ commits & 88.3\% \\
        Developers with $\geq 200$ commits & 70.6\% \\
        \hline
    \end{tabular}
\end{table}

Commit activity is concentrated in the later years of the Full Analysis Period: 2014 accounts for 1.2\% of commits, 2015 for 8.4\%, 2016 for 19.7\%, 2017 for 50.0\%, and January--March 2018 for 20.7\%. This pattern reflects the larger volume of observed activity in later years and the selection of developers who remain active through early 2018.

\begin{table}[htbp]
    \centering
    \small
    \caption{Weekly per-capita commit activity during the Full Analysis Period.}
    \label{tab:weekly_activity}
    \begin{tabular}{l|r}
        \hline
        Statistic & Weekly commits per capita \\
        \hline
        Mean & 1.8840 \\
        Standard deviation & 1.7150 \\
        Coefficient of variation & 0.9103 \\
        Median & 1.2509 \\
        Minimum & 0.1928 \\
        Maximum & 7.9133 \\
        \hline
    \end{tabular}
\end{table}

The aggregate coefficient of variation of 0.91 (Table~\ref{tab:weekly_activity}) indicates substantial week-to-week variation in community activity across the Full Analysis Period. The mean of 1.88 commits per developer per week is consistent with a cohort of persistent contributors whose activity nevertheless varies over time.

\begin{table}[htbp]
    \centering
    \small
    \caption{Developer activity during the Warmup ICL Period (January 22--February 18, 2018).}
    \label{tab:warmup_activity}
    \begin{tabular}{l|r}
        \hline
        Statistic & Value \\
        \hline
        Mean weekly commits per capita & 4.9528 \\
        Standard deviation & 2.1601 \\
        Coefficient of variation & 0.4361 \\
        Active developers & 927 (85.5\%) \\
        \hline
    \end{tabular}
\end{table}

Weekly activity during the Warmup ICL Period (Table~\ref{tab:warmup_activity}) is approximately 2.6 times higher and less variable than activity across the Full Analysis Period. This difference is expected because the sample-selection procedure favors developers who remain active near the beginning of the experimental timeline.

Developers also participate broadly across projects, with a mean of 50.1 repositories and a median of 38 repositories per developer across the Full Analysis Period. This cross-project exposure provides diverse behavioral histories for agent construction while preserving substantial variation in repository participation.

\begin{figure}[htbp]
    \centering
    \includegraphics[width=0.95\linewidth]{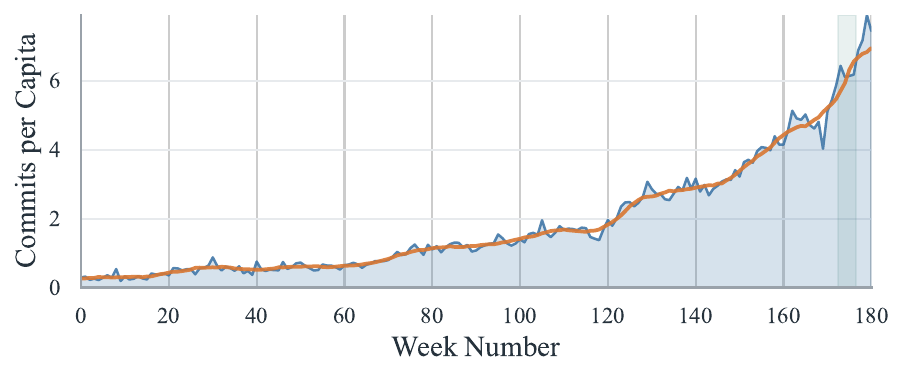}
    \caption{Weekly per-capita commit activity during the Full Analysis Period (October 1, 2014--March 18, 2018).}
    \label{fig:supp_weekly_timeseries}
\end{figure}

Figure~\ref{fig:supp_weekly_timeseries} plots the weekly per-capita commit time series across the Full Analysis Period, showing the secular upward trend in aggregate activity that underlies the distributional statistics reported above.

\paragraph{(3) Activity Stability Score.}

We measure individual temporal regularity across the Full Analysis Period using the Activity Stability Score (ASS):

\begin{equation}
\mathrm{ASS}_i=\frac{1}{1+\mathrm{CV}_i},
\end{equation}

where $\mathrm{CV}_i$ is the coefficient of variation in developer $i$'s weekly commit counts. Scores closer to 1 indicate regular weekly activity, whereas scores closer to 0 indicate irregular or bursty contribution patterns.

\begin{table}[htbp]
    \centering
    \small
    \caption{Distribution of Activity Stability Scores calculated over the Full Analysis Period. Scores are available for 1,082 of the 1,084 developers.}
    \label{tab:ass_distribution}
    \begin{tabular}{l|r}
        \hline
        Statistic & ASS \\
        \hline
        Mean & 0.4990 \\
        Median & 0.4986 \\
        Standard deviation & 0.0728 \\
        25th percentile & 0.4574 \\
        75th percentile & 0.5409 \\
        Minimum & 0.2363 \\
        Maximum & 1.0000 \\
        \hline
    \end{tabular}
\end{table}

\begin{figure}[htbp]
    \centering
    \includegraphics[width=0.95\linewidth]{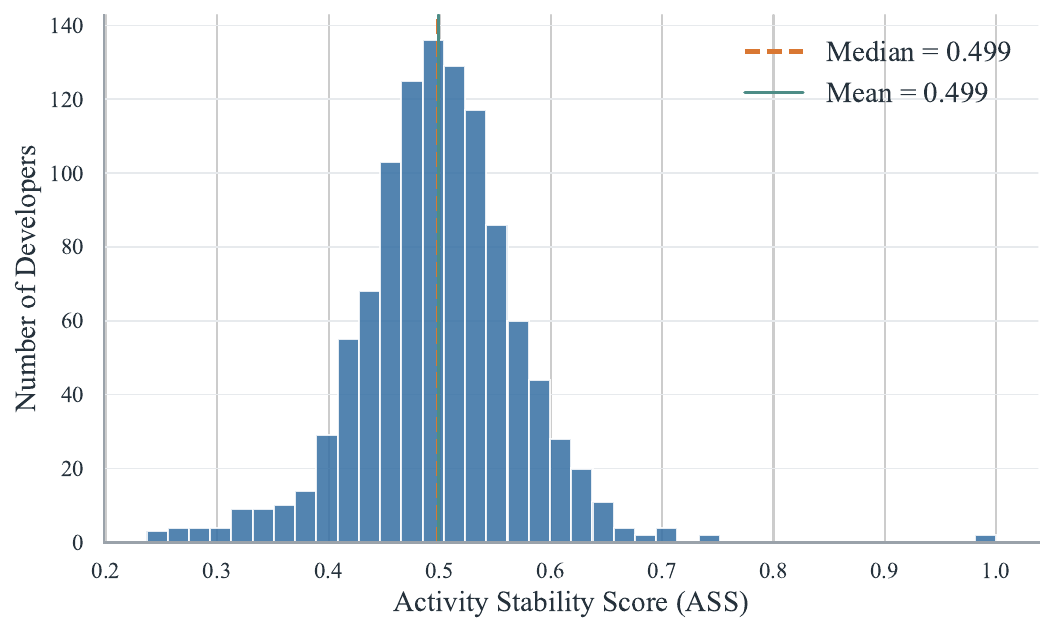}
    \caption{Distribution of Activity Stability Scores across 1,082 developers during the Full Analysis Period.}
    \label{fig:supp_ass_distribution}
\end{figure}

The distribution centers near 0.50 (Table~\ref{tab:ass_distribution}), indicating moderate temporal regularity at the individual level. Its full range preserves both highly bursty and highly regular developers, providing meaningful variation in activity rhythms for agent modeling. Figure~\ref{fig:supp_ass_distribution} visualizes this distribution.

\paragraph{(4) Repository-Based Collaboration Network.}
\label{sec:supp_network}

We construct the repository-based collaboration network from repository participation observed across the Full Analysis Period. Two sampled developers are connected when they contribute to at least one shared repository. Table~\ref{tab:network_statistics} reports the structural properties of this network, including density, degree, clustering, connected components, and the size of the largest component.

\begin{table}[htbp]
    \centering
    \small
    \caption{Structural properties of the repository-based developer network constructed from the Full Analysis Period.}
    \label{tab:network_statistics}
    \begin{tabular}{l|r}
        \hline
        Network statistic & Value \\
        \hline
        Number of nodes & 1,084 \\
        Number of edges & 2,963 \\
        Network density & 0.005048 \\
        Mean degree & 5.47 \\
        Median degree & 0 \\
        Maximum degree & 79 \\
        Number of connected components & 720 \\
        Share of nodes in the largest component & 26.0\% \\
        Average clustering coefficient & 0.2200 \\
        Number of isolated nodes & 677 (62.5\%) \\
        \hline
    \end{tabular}
\end{table}

The network is sparse and fragmented. Although developers participate in many repositories overall, most do not share a repository with another developer in the selected sample: the median degree is 0, and 677 developers are isolated. The mean degree of 5.47 is driven by a smaller group of well-connected developers, with a maximum degree of 79. The largest connected component contains 26.0\% of the sample, while the remaining developers are distributed across small components or isolated positions.

\begin{figure}[htbp]
    \centering
    \includegraphics[width=0.95\linewidth]{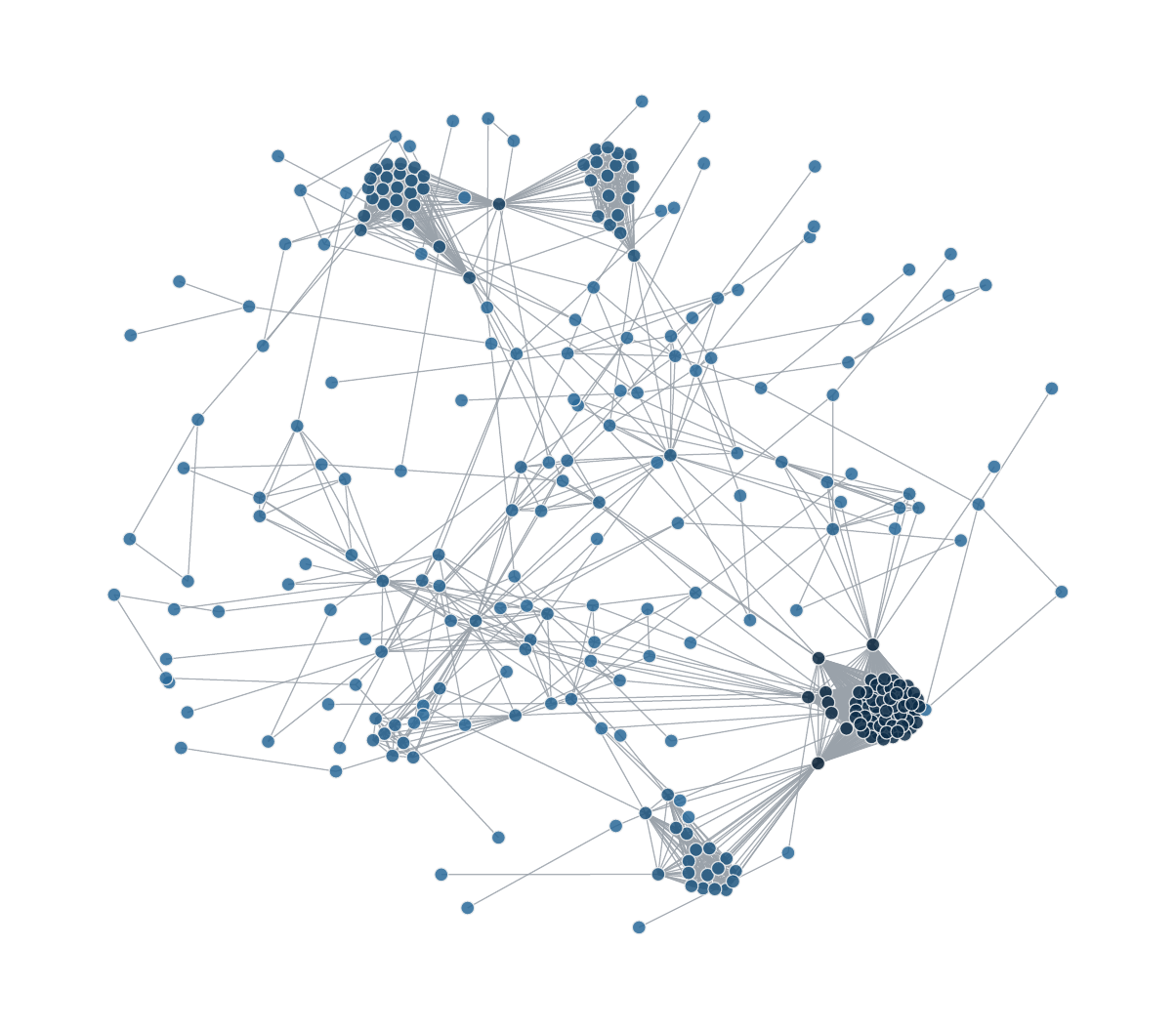}
    \caption{Repository-based collaboration network constructed from the Full Analysis Period for the 1,084 selected developers.}
    \label{fig:supp_network}
\end{figure}

Isolation refers only to the absence of ties to other developers in the analytical sample and does not imply that a developer has no collaborators outside the sample. The resulting structure combines a connected core with many peripheral developers, preserving meaningful variation in network position for modeling collaboration and CA diffusion. Figure~\ref{fig:supp_network} visualizes this network.

\paragraph{(5) Community Structure and Cross-Community Repositories.}
\label{sec:supp_crosscommunity}

Community detection is applied to the collaboration network constructed from the Full Analysis Period. It identifies nine developer communities within the connected portion of the network (Table~\ref{tab:crosscommunity}). Most collaboration remains localized within these communities, while 49 repositories connect developers from more than one community.

\begin{table}[htbp]
    \centering
    \small
    \caption{Community structure and cross-community repository connections during the Full Analysis Period.}
    \label{tab:crosscommunity}
    \begin{tabular}{l|r}
        \hline
        Statistic & Value \\
        \hline
        Number of detected communities & 9 \\
        Number of repositories shared across communities & 49 \\
        \hline
    \end{tabular}
\end{table}

This structure provides two pathways for community interaction: relatively concentrated collaboration within local groups and more limited exchange through cross-community repositories. Together, they form the real network through which repository discovery, interpersonal collaboration, and CA-related information can spread during the simulation.


\subsection{4. Developer-Agent Profile Initialization from Real Data}
\label{sec:supp_initialization}

Following the sample construction in Section~\ref{sec:supp_sample}, this section describes the \emph{Initialization} stage in which each selected developer's record is converted into a simulation-ready agent profile. The profile combines public background information, historical activity, repository participation, collaboration context, behavioral tendencies, and an operational persona for technology adoption. The purpose of this stage is to preserve observed differences in developer background, activity, repository participation, and network position before the Warmup and Simulation periods begin.

\paragraph{(1) Profile Composition.}
\label{sec:supp_data_schema}

Each developer profile is constructed from public GitHub records observed during the Historical Background Period and the Warmup ICL Period. The profile contains five components:

\begin{enumerate}
    \item \textbf{Public background information:} available profile fields such as biography, location, company, and follower relationships.
    \item \textbf{Technical and project experience:} programming-language coverage, major repositories, repository ownership, and participation across projects.
    \item \textbf{Activity and collaboration history:} contribution volume, active periods, weekly activity patterns, collaborator relationships, and network position.
    \item \textbf{Behavioral priors:} developer-specific tendencies toward common repository and contribution activities.
    \item \textbf{Biographical and adoption persona:} an evidence-based narrative of the developer's work patterns and an Innovation Diffusion Theory (IDT) persona used to represent differences in willingness to adopt new tools.
\end{enumerate}

These components provide the factual and behavioral context used by the developer agent throughout the simulation. Public records are used only when available, and missing fields are left unspecified rather than inferred. The resulting profiles therefore preserve real heterogeneity without requiring every developer to have the same amount or type of information.

\paragraph{(2) Quantitative Behavioral Profile.}
\label{sec:supp_action_probs}

The quantitative part of the profile summarizes each developer's historical contribution patterns. It includes total activity, active periods, weekly regularity, repository participation, programming-language coverage, and collaboration relationships. These measures describe both the scale of a developer's activity and the contexts in which that activity occurs.

We further estimate individual behavioral priors from real activities in the Warmup ICL Period. Activities are grouped into six broad categories: regular commits, repository creation, repository forking, issue opening, issue resolution, and issue discussion. For each developer, the prior for a category reflects how often that behavior appears across observed days.

Because some behaviors occur infrequently, raw individual rates may be unstable. We therefore shrink each estimate toward the corresponding population median. This procedure retains systematic differences among developers while reducing the influence of sparse observations. The resulting priors guide the relative likelihood of different activities during the Simulation Period.

\paragraph{(3) Biographical Narrative Construction.}
\label{sec:supp_biography}

The structured profile is supplemented with a concise biographical narrative that helps the LLM interpret the developer's real history as a coherent working background. The narrative is generated from two forms of evidence:

\begin{enumerate}
    \item \textbf{Structured summary:} contribution volume, active periods, programming-language use, repository participation, project roles, and collaboration patterns.
    \item \textbf{Historical activity records:} chronologically ordered repository contexts, commit messages, and timestamps from the Historical Background Period.
\end{enumerate}

Using these inputs, the LLM produces a second-person description of the developer's technical experience, recurring tasks, project responsibilities, and working habits. The prompt instructs the model to summarize only evidence supported by the real record and to avoid inventing personal details, motivations, or experiences. The complete prompt is provided in Section~\ref{sec:supp_prompts}(1).

The narrative does not replace the structured data. Instead, it provides a compact semantic interpretation that can be included in downstream prompts while the quantitative profile remains the source of factual activity and repository information.

\paragraph{(4) IDT-Based Adoption Persona.}
\label{sec:supp_idt}

To represent heterogeneity in coding agent adoption, each developer is assigned an operational persona based on Innovation Diffusion Theory (IDT)~\cite{Rogers2003Diffusion}. This persona is not treated as a direct psychological measurement of the real developer. Rather, it provides a transparent simulation mechanism for linking observed technical and collaborative histories to different levels of openness toward a new development tool.

Persona construction follows two stages. First, the LLM evaluates the real biography along four dimensions:

\begin{itemize}
    \item \textbf{Technology exploration:} the extent to which the developer works across new languages, frameworks, or technical domains.
    \item \textbf{Social influence:} the extent of collaboration, repository ownership, and participation across projects.
    \item \textbf{Risk willingness:} the extent to which the developer undertakes unfamiliar, experimental, or substantial technical changes.
    \item \textbf{Adoption pace:} the tendency to engage with new tools independently or only after wider use by peers.
\end{itemize}

Second, the four dimension scores are combined and mapped to five IDT categories according to their percentile positions in the sample distribution:

\begin{table*}[tbp]
    \centering
    \small
    \caption{Operational IDT persona categories used in the simulation.}
    \label{tab:idt_thresholds}
    \begin{tabular}{l|c|c}
        \hline
        IDT category & Percentile range & Simulated adoption orientation \\
        \hline
        Innovator & $\geq P_{95}$ & Highly open to experimenting with new tools \\
        Early Adopter & $P_{75}$--$P_{95}$ & Open to early use and attentive to emerging practices \\
        Early Majority & $P_{35}$--$P_{75}$ & Willing to adopt after observing credible benefits \\
        Late Majority & $P_{10}$--$P_{35}$ & Cautious and responsive to broader social evidence \\
        Laggard & $<P_{10}$ & Reluctant to adopt without strong practical need \\
        \hline
    \end{tabular}
\end{table*}

Using percentile thresholds preserves variation across the developer sample and avoids asking the LLM to assign adopter labels directly. The resulting persona is incorporated into the developer's self-description and informs later CA-related reasoning. It affects the agent's willingness to consider the coding agent but does not determine adoption on its own. Actual use remains a task-level decision shaped by awareness, peer activity, task characteristics, previous experience, and the current repository context. The complete prompt is provided in Section~\ref{sec:supp_prompts}(2).


\subsection{5. Agent Architecture}
\label{sec:supp_agent_arch_ch3}

Building on the developer profiles initialized in Section~\ref{sec:supp_initialization}, this section describes the multi-agent architecture through which developers, repositories, and the GitHub-like platform interact. It explains the roles of the developer and platform agents, the information retained in each developer's runtime state, and the daily Query--Act--Reflect workflow used to retrieve platform information, perform development activities, and update memory. This architecture provides a common execution environment for both the \emph{Warmup} and \emph{Simulation} periods, while the experimental protocol governing these periods is described separately in Section~\ref{sec:supp_sim_protocol}.

\paragraph{(1) Overall Multi-Agent Architecture.}
\label{sec:supp_agent_arch}

The simulation is implemented with AgentSociety~\cite{Piao2025AgentSociety}. We use its agent lifecycle, memory, and message-passing infrastructure while replacing its original urban-simulation functions with domain-specific mechanisms for software development and repository collaboration.

The simulated community contains two types of agents. Each selected developer is represented by a developer agent initialized from the real profile described in Section~\ref{sec:supp_initialization}. A platform agent represents the shared GitHub environment and maintains repositories, commits, issues, participation relationships, and public activity records. Developer agents make development decisions independently, whereas the platform agent coordinates shared information and records the outcomes of those decisions.

Each simulation day is organized into three operational phases. Developer agents first request updated platform information, then wait for the platform to process and return the relevant records, and finally make decisions, perform activities, and update their internal states. This separation ensures that daily decisions are based on a consistent platform state.

\paragraph{(2) Developer Agent State.}
\label{sec:supp_agent_state}

Each developer agent maintains a runtime state that combines relatively stable profile information with information that changes during the simulation. The stable component includes the developer's technical background, repository experience, collaboration relationships, behavioral priors, and adoption persona. The dynamic component includes recent platform observations, pending tasks, activity history, issue conversations, long-term memory, and current awareness or use of the coding agent.

This state serves three purposes. First, it preserves continuity across simulation days by retaining previous activities and unresolved tasks. Second, it provides the context required for repository and task decisions. Third, it records changes in the developer's experience, collaboration history, and relationship with the coding agent. The same types of information are maintained for all agents, although their values differ according to each developer's real profile and simulated experience.

\paragraph{(3) GitHub Platform Agent.}
\label{sec:supp_platform}

The platform agent provides the shared infrastructure through which developer activities become visible to the simulated community. It records repository creation and forking, commits, issues, developer participation, and interaction histories, but does not decide which activities developers should perform.

At the beginning of each simulated day, the platform returns recent commits and issues from repositories owned by or involving the requesting developer. It also recommends repositories through the existing collaboration network by identifying projects involving the developer's collaborators but not yet involving the developer. These recommendations represent network-based repository discovery rather than content-based recommendation.

The platform also supports the interaction function between developers. For example, it delivers issue responses to the original issue opener and makes CA-assisted commits visible to collaborators. These shared records provide the public information through which collaboration, repository discovery, and CA awareness can develop during the simulation.

\subsection{6. Query--Act--Reflect Daily Workflow}
\label{sec:supp_daily_workflow}

Each developer agent follows a daily \emph{Query--Act--Reflect} workflow:

\begin{itemize}
    \item \textbf{Query:} The agent observes the current platform environment and retrieves the information needed for daily decisions.
    \item \textbf{Act:} The agent selects and performs development activities under task and time constraints.
    \item \textbf{Reflect:} The agent interprets the day's observations and outcomes and updates its memory and behavioral tendencies.
\end{itemize}

The complete prompts supporting these stages are provided in Section~\ref{sec:supp_prompts}.

\paragraph{(1) Query: Retrieving Platform and Task Context.}
\label{sec:supp_query}

At the beginning of each day, the agent retrieves recent commits, issues, repository updates, pending work, and network-based repository recommendations. It combines this information with its initialized profile, current repository responsibilities, behavioral priors, task history, and long-term memory. The resulting context allows the agent to respond to both stable individual characteristics and recent changes in the community.

The Query stage also supports social observation. Developers can observe public activities in shared repositories, including CA-assisted contributions made by collaborators. Such information may later affect their awareness of the coding agent and their willingness to consider its use.

The complete prompt is provided in Section~\ref{sec:supp_prompts}(3).

\paragraph{(2) Act: Behavior Selection.}
\label{sec:supp_action_selection}

The Act stage begins with routine responsibilities, such as reviewing collaborators' commits and responding to issues in repositories owned by the developer. The agent may then perform additional activities, including maintaining owned repositories, contributing to repositories owned by others, opening issues, creating repositories, and forking repositories discovered through the collaboration network.

Daily behavior selection combines a real activity baseline with the agent's current context. Developers are assigned different baseline activity levels according to their historical behavior, while recent activity and pending work can adjust the amount of work considered for the current day. The agent then selects a limited set of activities using its individual behavioral priors. This process preserves stable differences across developers without making their daily behavior fully deterministic. The complete prompts for creating repositories, forking repositories, and opening issues are provided in Section~\ref{sec:supp_prompts}(4), (5), and (6), respectively.

\paragraph{(3) Act: Repository Maintenance and Task Execution.}
\label{sec:supp_task_lifecycle}

When an agent decides to maintain a repository, the process follows three stages:

\begin{enumerate}
    \item \textbf{Repository selection:} The agent identifies repositories that require attention based on project descriptions, participation roles, recent activity, and existing tasks.
    \item \textbf{Task planning:} The agent generates specific and non-duplicate tasks that are consistent with the repository's purpose and current state.
    \item \textbf{Task execution:} The agent selects tasks that can be completed within the available daily time budget and produces the corresponding commit records.
\end{enumerate}

Unfinished tasks remain available for later days, allowing project work to continue across multiple decision cycles. Recent patterns of sustained activity or rest may adjust the amount of work that an agent can complete, representing short-term variation in working capacity. The complete prompts for task generation, repository selection, and task execution are provided in Section~\ref{sec:supp_prompts}(7), (8), (9), (10), and (11).

Under the CA condition, developers who are aware of the coding agent may decide whether to use it for a specific task. The coding agent can assist with code understanding, generation, debugging, testing, and revision, but the developer agent retains control over task selection and final submission. CA use is therefore modeled as a task-level choice rather than a permanent replacement of human decision-making.

\paragraph{(4) Act: Cross-Developer Interaction.}
\label{sec:supp_collab_design}

Cross-developer interaction occurs mainly through commit review and issue handling. When a collaborator contributes to a developer's repository, the repository owner reviews the contribution and may use the coding agent as support. To keep the model focused on collaboration pathways rather than software-integration failures, submitted commits are treated as accepted after review. Rejection, rollback, and merge conflicts are not modeled.

For issues opened by other developers, the repository owner may produce one of three outcomes:
\begin{itemize}
    \item \textbf{Fully resolved:} The issue is completely addressed and produces a corresponding commit.
    \item \textbf{Partially resolved:} Part of the issue is addressed only in part, after which the owner explains the remaining problem or requests further information.
    \item \textbf{Not solved:} The issue cannot be resolved at that time, and the owner explains the reason or identifies the additional information or work required.
\end{itemize}

The platform forwards these responses to the issue opener and preserves the conversation history. Agents can consult earlier exchanges when making later decisions, allowing unresolved or partially resolved issues to progress as new information becomes available.

These mechanisms produce visible interpersonal records through which developers exchange technical information. When CA support is used, the resulting activity remains associated with the participating developers and repositories, allowing the simulation to distinguish direct human interaction from CA-assisted collaboration. The complete prompts for commit review and issue resolution are provided in Section~\ref{sec:supp_prompts}(12), (13), and (14).

\paragraph{(5) Reflect: Updating Memory and Behavioral Tendencies.}
\label{sec:supp_memory_design}

The Reflect stage updates two complementary forms of memory. 
\begin{itemize}
    \item \textbf{Historical activities} consist of deterministically maintained statistics, including cumulative commits, active days, weekly activity, major repositories, and recent activity levels. 
    \item \textbf{Long-term memory} is a concise LLM-maintained narrative that summarizes technical interests, project progress, collaboration experiences, and awareness of the coding agent. 
\end{itemize}
Keeping these components separate preserves factual accuracy while allowing the agent to interpret its experience in context.

Reflection occurs through two pathways. Environmental reflection processes important platform information, including repository changes, collaborator activity, and CA-related signals. Post-action reflection summarizes the activities performed during the day and their outcomes. Together, these processes update the agent's memory and future behavioral orientation while keeping factual activity statistics separate from LLM-generated interpretation. The complete prompts for environmental reflection, post-action reflection, and tendency updates are provided in Section~\ref{sec:supp_prompts}(15), (16), (17), and (18).

At the end of each day, the updated state and activity events are stored for use in subsequent decisions and later analysis. This daily persistence allows the simulation to represent cumulative experience, continuing tasks, repeated interaction, and the gradual diffusion and adoption of the coding agent.


\subsection{7. Warmup and Simulation Procedure}
\label{sec:supp_sim_protocol}

Following agent initialization and the specification of the Query--Act--Reflect workflow, this section describes how the simulation is conducted across two consecutive phases: the \emph{Warmup ICL Period} and the \emph{Simulation Period}. The Warmup phase guides agents to learn their corresponding real activity patterns, whereas the Simulation phase allows agents to act autonomously under the No-CA and CA conditions. Our experimental design ensures that both simulation conditions begin from the same real-data evolved community state and differ only in whether the coding agent is introduced.

\paragraph{(1) Experimental Timeline.}
\label{sec:supp_timeline}
The experiment consists of two consecutive phases:
\begin{itemize}
    \item \textbf{Warmup ICL Period} (January 22--February 18, 2018, 4 weeks): Real GitHub activities are replayed to update developer memory, activity history, repository states, and collaboration context.
    \item \textbf{Simulation Period} (February 19--March 18, 2018, 4 weeks): Real activities are no longer replayed. Developer agents operate autonomously through the Query--Act--Reflect workflow under either the No-CA or CA condition.
\end{itemize}

The Warmup phase provides a shared real starting point, while the Simulation phase generates the two community evolution trajectories used in the analysis.

\paragraph{(2) Warmup ICL Period.}
\label{sec:supp_icl}

During the Warmup ICL Period, the system provides each developer agent with the activities and task list recorded during the corresponding four-week period. These records include repository contributions, repository creation, and issue-related activities, and are assigned to owned or participated repositories according to the developer's observed relationship with each project.

Agents operate through the Query--Act--Reflect workflow described in Section~\ref{sec:supp_daily_workflow}. They retrieve platform information, carry out the provided activities, interact with other developers, observe the resulting repository changes, and update their memory and behavioral tendencies. Only activity and task selection are predefined: instead of deciding independently what to do, each agent follows the sequence derived from the developer's real activity record. Processing these activities through the complete daily workflow allows the agent to recognize the developer's recent activity patterns, repository routines, and collaboration habits from in-context examples. Warmup therefore supports in-context adaptation without training the model or changing the parameters of the underlying LLM.

Throughout the Warmup Period, the platform updates shared repository, commit, issue, participation, and interaction states. The Coding Agent is not introduced at this stage, and no developer receives CA-related information or uses CA assistance. CA awareness and adoption therefore begin only after the intervention in the Simulation Period.

\paragraph{(3) Shared Post-Warmup State.}
\label{sec:supp_branching}

At the end of Warmup, the system saves a common community state containing the updated developer profiles and memories, repository and task states, contribution histories, participation relationships, collaboration network, and platform information.

The No-CA and CA conditions both begin from this same post-Warmup state. They therefore share the same developers, repositories, recent real histories, network structures, and behavioral mechanisms. This branching design isolates the coding agent intervention as the main difference between the two experimental conditions.

\paragraph{(4) Simulation Period.}
\label{sec:supp_simulation_period}

During the Simulation period, agents no longer receive preloaded real activities. Each developer autonomously retrieves platform information, selects and performs development activities, interacts with other developers, and updates memory through the Query--Act--Reflect workflow described in Section~\ref{sec:supp_daily_workflow} as well.

Both No-CA and CA conditions use the same simulation length, initialized community, platform mechanisms, activity constraints, and task processes. They differ only in whether the coding agent is available.  The coding agent can support code understanding, generation, debugging, testing, and revision, while the developer agent retains control over task selection and final submission.

\textbf{No-CA condition.}
In the No-CA condition, the coding agent is not introduced. Developers complete repository maintenance, task execution, commit review, and issue handling without CA assistance, and the platform provides no CA-related information. This condition represents the trajectory of the community in the absence of the new tool and serves as the baseline for comparison.

\textbf{CA condition.}
In the CA condition, the coding agent is introduced at the beginning of the Simulation Period. The intervention and subsequent adoption process follow three stages:

\begin{enumerate}
   \item \textbf{Initial intervention:} The platform announces the coding agent to a group of seed developers selected for their broad technical experience and high recent activity. Seed developers are selected according to technical coverage and Warmup activity. Technical coverage is identified from the names and descriptions of repositories owned by or involving each developer. The 14 technical domains are Python, JavaScript, TypeScript, Go, Rust, Java, C/C++/C\#, Ruby, PHP, Swift/iOS, Shell, Kotlin, Dart/Flutter, and R. A seed developer must cover at least seven technical domains and complete at least 15 commits during Warmup. These developers become aware of the tool but retain the choice of whether to use it.
    \item \textbf{Social diffusion:} CA-assisted activities remain visible in public repository records. Other developers can learn about the coding agent by observing CA-assisted activities produced by collaborators in shared repositories. Awareness therefore spreads through the existing developer--repository network rather than through a community-wide announcement.
    \item \textbf{Task-level adoption:} Once aware of the coding agent, a developer may decide whether to use it for a specific development task, commit review, or issue-resolution activity. Adoption does not imply permanent or universal use. Developers' willingness to use the coding agent may change as they receive external information, observe collaborators, and gain direct experience with the tool. Each decision remains dependent on the task, developer profile, previous experience, and current repository context.
\end{enumerate}

\paragraph{(5) Repeated Runs and Randomness Control.}
\label{sec:supp_repetition}

Each experimental condition is run independently three times to account for variation in LLM-generated decisions. All runs use the same real sample, post-Warmup state, simulation timeline, and experimental configuration. Initialization and static community structures are held constant, while stochastic variation in agent decisions is retained.

Results are calculated separately for each run and then aggregated within each condition. This design preserves the variability of generative-agent behavior while supporting consistent comparisons between the No-CA and CA trajectories.


\begin{table*}[htbp]
    \centering
    \scriptsize
    \caption{Full Prompt Set Used in this study.}
    \label{tab:prompt_inventory}
    \setlength{\tabcolsep}{2pt}
    \renewcommand{\arraystretch}{0.9}
    \begin{tabular}{p{0.45\linewidth}|p{0.14\linewidth}|p{0.34\linewidth}}
        \hline
        Prompt name & Phase & Function \\
        \hline
        Biography Construction Prompt & Initialization & Construct a developer's biographical narrative \\
        IDT Classification Prompt & Initialization & Score behavioral dimensions used to assign an IDT persona \\
        ACTION\_SELECTION\_PROMPT & Query & Select today's GitHub behavior subset \\
        CREATE\_REPO\_PROMPT & Act & Decide whether to create a new repository \\
        FORK\_REPO\_PROMPT & Act & Decide whether to fork a collaborator's repository \\
        ISSUE\_MESSAGE\_PROMPT & Act & Decide whether to open an issue and generate content \\
        GENERATE\_TASKS\_PROMPT & Act & Generate pending tasks for a repository \\
        SELECT\_REPOS\_FOR\_MAINTENANCE\_PROMPT & Act & Select repositories needing new tasks \\
        CHECK\_REPO\_FOR\_MAINTENANCE\_PROMPT & Act & Check whether a single repository needs maintenance \\
        GLOBAL\_TASK\_COMMIT\_SELECTION\_PROMPT\_WITH\_CA & Act & Select tasks globally with CA fields \\
        GLOBAL\_TASK\_COMMIT\_SELECTION\_PROMPT\_WITHOUT\_CA & Act & Select tasks globally without CA fields \\
        OWNED\_REPO\_OTHERS\_COMMIT\_REVIEW\_PROMPT & Act & Review collaborators' commits \\
        OWNED\_REPO\_OTHERS\_ISSUE\_RESOLUTION\_PROMPT\_WITH\_CA & Act & Resolve collaborators' issues with CA fields \\
        OWNED\_REPO\_OTHERS\_ISSUE\_RESOLUTION\_PROMPT\_WITHOUT\_CA & Act & Resolve collaborators' issues without CA fields \\
        ENVIRONMENT\_MEMORY\_UPDATE\_PROMPT & Reflect & Update long-term memory from news \\
        MULTI\_BEHAVIOUR\_MEMORY\_PROMPT & Reflect & Update memory from daily actions \\
        TENDENCY\_UPDATED\_PROMPT\_WITH\_NEWS & Reflect & Update CA usage tendency from external information \\
        TENDENCY\_UPDATED\_PROMPT\_WITH\_RECORDS & Reflect & Update CA usage tendency from direct experience \\
        \hline
    \end{tabular}
\end{table*}

\newtcolorbox{promptbox}{
  colback=gray!5,
  colframe=gray!50,
  sharp corners,
  boxrule=0pt,
  leftrule=3pt,
  enhanced,
  breakable,
  fontupper=\scriptsize
}

\subsection{8. Full Prompt Set}
\label{sec:supp_prompts}

This section presents the full set of prompts used in the study. Two prompts support offline developer-profile initialization, and 16 support the daily simulation workflow. Table~\ref{tab:prompt_inventory} summarizes each prompt by its name, workflow stage, and function, followed by the full prompt texts.

\textbf{(1) Biography construction prompt.}
\label{sec:supp_prompt_profile}
This prompt combines role instructions with a developer's historical commit records to generate a concise and factual biography. The resulting narrative summarizes changes in technical focus, recurring project responsibilities, and working habits while avoiding claims that are not supported by the input data.

\begin{promptbox}
You write concise, reflective summaries of a person's past behavior.
Write in second person (use 'you'). Do not invent facts.

Below is a list of commit events from the same user, spanning multiple years.
Write a short narrative summary in second person describing how this person's development focus and habits have evolved over the years.

Guidelines:
\begin{itemize}[label=-, leftmargin=1.5em, nosep]
\item Use second person ('you') throughout
\item Focus on long-term patterns and shifts
\item Do NOT list commits or repeat timestamps
\item If data is insufficient, state the uncertainty
\item Limit the response to no more than 300 words
\end{itemize}

Commit events:
\{commit\_text\}
\end{promptbox}

\textbf{(2) IDT Classification Prompt.}
\label{sec:supp_prompt_idt}
This prompt converts the background story into four bounded behavioral scores and a brief justification for subsequent IDT classification.

\begin{promptbox}
You are analyzing a GitHub developer's behavioral profile based on their activity history.
Rate this developer on the following 4 dimensions, each on a 1-5 scale (1=lowest, 5=highest):

1. Technology Exploration (1=pure maintenance of familiar tools, 5=constantly trying new languages, frameworks, or experimental stacks)
\begin{itemize}[label=-, leftmargin=1.5em, nosep]
\item Clues: diversity of programming languages in repos, side projects exploring unfamiliar domains, early adoption of bleeding-edge tools
\end{itemize}

2. Social Influence (1=isolated, works alone on personal repos, 5=highly connected, collaborates across many repos/projects)
\begin{itemize}[label=-, leftmargin=1.5em, nosep]
\item Clues: contributing to many different projects, owning repos with external contributors, interacting with diverse developers
\end{itemize}

3. Risk Willingness (1=conservative, sticks to established patterns, 5=bold, frequently undertakes radical refactoring or unconventional approaches)
\begin{itemize}[label=-, leftmargin=1.5em, nosep]
\item Clues: rewriting codebases, trying unproven technologies, building from scratch instead of using established solutions
\end{itemize}

4. Adoption Pace (1=slow follower, waits until technology is mainstream, 5=independent early adopter who acts before peers)
\begin{itemize}[label=-, leftmargin=1.5em, nosep]
\item Clues: how early they adopt new tools/patterns relative to when they appear in the ecosystem, self-directed exploration vs following tutorials
\end{itemize}

Background story:
\{background\_story\}

Your output must be strictly valid JSON in the following format:
\{\{"technology\_exploration": <1-5>, "social\_influence": <1-5>, "risk\_willingness": <1-5>, "adoption\_pace": <1-5>, "summary": "<one sentence justifying the scores>"\}\}
\end{promptbox}

\textbf{(3) ACTION\_SELECTION\_PROMPT.}
\label{sec:supp_prompt_action_selection}
This Query-phase prompt selects the developer's activities for the day from five candidate behaviors using profile, memory, recent activity, and repository recommendations. It returns the selected subset and a brief reason. \texttt{issue\_resolution} remains active by default.

\begin{promptbox}
You are simulating this GitHub user in 2018.
Decide what this user would realistically want to do on GitHub today.

Profile:
\{background\}

Recent tech news:
\{news\}

General behavioral style:
\begin{itemize}[label=-, leftmargin=1.5em, nosep]
\item Your decisions should reflect your own background, interests, habits, and current attention.
\item You may choose any subset of the available behaviors.
\item Only choose behaviors that you genuinely feel motivated to do today.
\end{itemize}

Long-term memory:
\{long\_term\_memory\}

Recent commits in repositories owned by you:
\{owned\_repo\_commits\}

Recent commits in repositories you contributed to but do not own:
\{participated\_repo\_commits\}

Recommended repositories from collaborators or the platform:
\{recommended\_repos\}

Available behaviors:
\begin{itemize}[label=-, leftmargin=1.5em, nosep]
\item "work\_on\_owned\_repo\_tasks":
  Work on concrete tasks in one or more repositories you own.
  Choose this only when you are willing to enter the task-planning or task-completion phase for owned repositories today.
  This may include adding new pending tasks, updating existing pending tasks, completing pending tasks, or making commits based on completed tasks.
\item "work\_on\_participated\_repo\_tasks":
  Work on concrete tasks in one or more repositories you contribute to but do not own.
  Choose this only when you are willing to enter the task-planning or task-completion phase for participated repositories today.
  This may include adding new pending tasks, updating existing pending tasks, completing pending tasks, or making commits based on completed tasks.
\item "issues\_opened":
  Open an issue in a repository you already participate in.
  This behavior represents raising a problem, proposing a task, asking for coordination, or starting a discussion around a concrete issue.
\item "repos\_created":
  Create a brand new repository of your own.
  This behavior represents starting a new project direction, creating a new technical space for work, or setting up a repository for an idea not currently covered by your existing repositories.
\item "repos\_forked":
  Fork one recommended repository owned by a collaborator that you do not already participate in.
  This behavior represents deciding that a repository is relevant enough to your interests or future work that you want your own copy.
\end{itemize}

Decision principle:
\begin{itemize}[label=-, leftmargin=1.5em, nosep]
\item Select behaviors based on what feels natural and plausible for this user today.
\item Use both long-term memory and short-term memory to decide what stands out.
\end{itemize}

Return ONLY one valid JSON object with exactly these fields:
\begin{itemize}[label=-, leftmargin=1.5em, nosep]
\item "selected\_actions": a JSON array containing zero or more behaviors from the list above
\item "decision\_reason": a short explanation within 50 words of why these actions were chosen today
\end{itemize}

Do not return any extra text outside the JSON object.
\end{promptbox}

\textbf{(4) CREATE\_REPO\_PROMPT.}
\label{sec:supp_prompt_create_repo}
This prompt decides whether to create a repository and, if so, generates its name and description.

\begin{promptbox}
You are simulating this GitHub user in 2018.
Decide whether this user would create a new GitHub repository today.

Profile:
\{background\}

Long-term memory:
\{long\_term\_memory\}

Recent tech news:
\{news\}

Current owned repositories:
\{owned\_repo\_list\}

Current participated repositories:
\{participated\_repo\_list\}

Decide whether creating a new repository feels natural and worthwhile for this user today.

A new repository is appropriate when:
\begin{itemize}[label=-, leftmargin=1.5em, nosep]
\item you want to start a genuinely new project direction
\item the idea is not already well covered by your existing repositories
\item today's context gives you a plausible reason to start something new
\item only create a repository if there is a clear motivation.
\end{itemize}

Do not create a new repository when:
\begin{itemize}[label=-, leftmargin=1.5em, nosep]
\item your current repositories already provide a good place for the work
\item there is no concrete or meaningful new direction worth separating into its own repository
\end{itemize}

Return ONLY one valid JSON object with exactly these fields:
\begin{itemize}[label=-, leftmargin=1.5em, nosep]
\item "repos\_created": "YES" or "NO"
\item "repo\_name": string
\item "repo\_description": string
\item "decision\_reason": a short explanation within 50 words
\end{itemize}
\end{promptbox}
\textbf{(5) FORK\_REPO\_PROMPT.}
\label{sec:supp_prompt_fork_repo}
This prompt decides whether to fork at most one recommended repository owned by a collaborator.

\begin{promptbox}
You are simulating this GitHub user in 2018.
Decide whether this user would fork a collaborator's repository today.

Profile:
\{background\}

Long-term memory:
\{long\_term\_memory\}

Recent tech news:
\{news\}

Candidate repositories owned by collaborators that you do not already participate in:
\{repo\_list\}

Decide whether forking one candidate repository feels natural and worthwhile for this user today.

A fork is appropriate when:
\begin{itemize}[label=-, leftmargin=1.5em, nosep]
\item a candidate repository is strongly relevant to your interests, ongoing work, or future plans
\item you plausibly want your own copy for modification, experimentation, or future contribution
\end{itemize}

Do not fork when:
\begin{itemize}[label=-, leftmargin=1.5em, nosep]
\item none of the candidates feel clearly relevant enough
\item the repositories are only mildly interesting but not worth adopting into your own workflow
\end{itemize}

You may fork at most one repository in this round.

If you decide not to fork, return:
\begin{itemize}[label=-, leftmargin=1.5em, nosep]
\item "repos\_forked": "NO"
\item "repo\_id": -1
\end{itemize}

If you decide to fork, return:
\begin{itemize}[label=-, leftmargin=1.5em, nosep]
\item "repos\_forked": "YES"
\item a "repo\_id" copied from the candidate list
\end{itemize}

Return ONLY one valid JSON object with exactly these fields:
\begin{itemize}[label=-, leftmargin=1.5em, nosep]
\item "repos\_forked": "YES" or "NO"
\item "repo\_id": integer
\item "decision\_reason": a short explanation within 50 words
\end{itemize}
\end{promptbox}
\textbf{(6) ISSUE\_MESSAGE\_PROMPT.}
\label{sec:supp_prompt_issue_message}
This prompt selects participated repositories that require coordination and generates one issue message for each selection.

\begin{promptbox}
You are simulating this GitHub user in 2018.
Decide whether participated repositories need issues, and write corresponding issue messages.

\begin{itemize}[label=-, leftmargin=1.5em, nosep]
\item You may open issues in zero or multiple repositories.
\item Only open an issue when there is a clear coordination need, or unresolved problem worth discussing.
\item Every selected repository must come from the candidate repository list.
\item For each selected repository, write one concise issue message.
\item Do not open more than one issue for the same repository.
\end{itemize}

Profile:
\{background\}

Past memory:
\{past\_memory\}

Tech news:
\{news\}

Candidate repositories:
\{repo\_list\}

Return ONLY a JSON array. Each item must include:
\begin{itemize}[label=-, leftmargin=1.5em, nosep]
\item "repo\_id": integer repository id from the candidate repository list
\item "message": one concise issue message for that repository
\end{itemize}

If no issue should be opened, return [].
If issues should be opened, return an array with one or more items.
\end{promptbox}
\textbf{(7) GENERATE\_TASKS\_PROMPT.}
\label{sec:supp_prompt_generate_tasks}
This prompt generates one to three actionable, non-duplicate tasks aligned with the repository context and developer profile.

\begin{promptbox}
You are simulating this GitHub user in 2018.
Generate plausible pending tasks for one GitHub repository.

Profile:
\{background\}

Long-term memory:
\{long\_term\_memory\}

Recent tech news:
\{news\}

Repository information:
ID: \{repo\_id\}
Name: \{repo\_name\}
Description:
\{repo\_description\}

Repository relation:
\{repo\_relation\}

Commits in the last 24 hours:
\{recent\_commits\}

Current Pending Tasks:
\{current\_tasks\}

Generate 1\textasciitilde{}3 short, concrete pending tasks that would be plausible for this user in this repository.

Requirements:
\begin{itemize}[label=-, leftmargin=1.5em, nosep]
\item Tasks must be consistent with the repository's purpose, current state, and recent updates.
\item Tasks must be consistent with the user's profile and long-term memory.
\item Tasks should look like realistic GitHub repository tasks, not vague goals or broad project plans.
\item Avoid duplicating or trivially rewording items already listed in Current Pending Tasks.
\item Keep each task short and actionable.
\end{itemize}

Return ONLY one valid JSON object with exactly this field:
\begin{itemize}[label=-, leftmargin=1.5em, nosep]
\item "new\_tasks": a JSON array of 1 to 3 task strings
\end{itemize}

Do not wrap the JSON in markdown or code fences.
\end{promptbox}
\textbf{(8) SELECT\_REPOS\_FOR\_MAINTENANCE\_PROMPT.}
\label{sec:supp_prompt_select_repos}
This prompt identifies which candidate repositories warrant new tasks, balancing current backlog against recent activity and repository relevance.

\begin{promptbox}
You are simulating this GitHub user in 2018.

Profile:
\{background\}

Long-term memory:
\{long\_term\_memory\}

Recent tech news:
\{news\}

Repository relation:
\{repo\_relation\}

Candidate repositories:
\{repo\_candidates\}

Select repositories to add NEW pending tasks to today.

Decision meaning:
\begin{itemize}[label=-, leftmargin=1.5em, nosep]
\item "YES": add new pending tasks today.
\item "NO": do not add new tasks today, even if existing pending tasks could be completed.
\end{itemize}

Rules:
\begin{itemize}[label=-, leftmargin=1.5em, nosep]
\item Base the decision only on the provided repository information, relation, tech news, recent commits, and current pending tasks.
\item Do not answer "YES" merely because the repository is important, active, or already has many pending tasks.
\item Current Pending Tasks represent backlog pressure.
\item Before CodeAgent is available or known, high backlog should reduce willingness to add new tasks.
\item After CodeAgent is available and known, moderate backlog may be less discouraging for small, well-scoped tasks.
\item Choose "YES" only when recent activity, repo context, or user interest clearly justifies adding new tasks today.
\item Otherwise choose "NO".
\end{itemize}

Return ONLY a JSON array.
Each item in the array must contain exactly these fields:
\begin{itemize}[label=-, leftmargin=1.5em, nosep]
\item "repo\_id": integer
\item "decision": "YES" or "NO"
\item "reason": a short explanation within 50 words
\end{itemize}
\end{promptbox}

\textbf{(9) CHECK\_REPO\_FOR\_MAINTENANCE\_PROMPT.}
\label{sec:supp_prompt_check_repo}
This prompt checks whether one repository warrants new pending tasks.

\begin{promptbox}
You are simulating this GitHub user in 2018.
Decide whether this user would actively maintain the given repository today.

Profile:
\{background\}

Long-term memory:
\{long\_term\_memory\}

Recent tech news:
\{news\}

Repository relation:
\{repo\_relation\}

Given repository:
\{repo\_info\}

Decide whether to add NEW pending tasks to this repository today.

Decision meaning:
\begin{itemize}[label=-, leftmargin=1.5em, nosep]
\item "YES": add new pending tasks today.
\item "NO": do not add new tasks today, even if existing pending tasks could be completed.
\end{itemize}

Rules:
\begin{itemize}[label=-, leftmargin=1.5em, nosep]
\item Base the decision only on the provided repository information, relation, tech news, recent commits, and current pending tasks.
\item Do not answer "YES" merely because the repository is important, active, or already has many pending tasks.
\item Current Pending Tasks represent backlog pressure.
\item Before CodeAgent is available or known, high backlog should reduce willingness to add new tasks.
\item After CodeAgent is available and known, moderate backlog may be less discouraging for small, well-scoped tasks.
\item Choose "YES" only when recent activity, repo context, or user interest clearly justifies adding new tasks today.
\item Otherwise choose "NO".
\end{itemize}

Return ONLY a JSON object.
The object must contain exactly these fields:
\begin{itemize}[label=-, leftmargin=1.5em, nosep]
\item "decision": "YES" or "NO"
\item "reason": a short explanation within 50 words
\end{itemize}
\end{promptbox}

\textbf{(10) GLOBAL\_TASK\_COMMIT\_SELECTION\_\allowbreak{}PROMPT\_WITH\_CA.}
\label{sec:supp_prompt_global_task_ca}
This CA-aware variant additionally records CA use and speedup while preserving the shared daily time budget.

\begin{promptbox}
You are simulating this GitHub user in 2018.
Decide which pending tasks to complete today across ALL your repositories.

Profile:
\{background\}

Long-term memory:
\{long\_term\_memory\}

Recent tech news:
\{news\}

Today you have at most 480 minutes total to complete tasks across all repositories.

Below are all your repositories that currently have pending tasks:
\{repo\_blocks\}

Decide which tasks, if any, to complete today.
You may select tasks from one or multiple repositories.

Rules:
\begin{itemize}[label=-, leftmargin=1.5em, nosep]
\item Select zero or more tasks from the Pending Tasks listed above.
\item Copy each selected task's text exactly as shown.
\item For each selected task, write one concise and plausible Git commit message.
\item For each selected task, decide whether to use CodeAgent based on your CodeAgent orientation, the task, and the repository context.
\item Estimate "time\_cost" as the actual completion time today after applying the selected CodeAgent decision, in integer minutes.
\item The total "time\_cost" across ALL selected tasks must not exceed 480 minutes.
\item If "codeagent\_decision" is "YES", estimate "codeagent\_speedup" as the acceleration factor provided by CodeAgent, greater than or equal to 1.0.
\item If "codeagent\_decision" is "NO", set "codeagent\_speedup" to 1.0.
\item Return an empty array if no task should be completed today.
\end{itemize}

Return ONLY a JSON array.
Each item in the array must contain exactly these fields:
\begin{itemize}[label=-, leftmargin=1.5em, nosep]
\item "repo\_id": integer repository ID
\item "task": one task string copied exactly from Pending Tasks
\item "commit\_message": one concise commit message for that task
\item "codeagent\_decision": "YES" or "NO"
\item "decision\_reason": a short explanation within 50 words
\item "time\_cost": the estimated actual completion time for that task in minutes as an integer
\item "codeagent\_speedup": the estimated acceleration factor provided by CodeAgent, as a number greater than or equal to 1.0
\end{itemize}
\end{promptbox}
\textbf{(11) GLOBAL\_TASK\_COMMIT\_SELECTION\_\allowbreak{}PROMPT\_WITHOUT\_CA.}
\label{sec:supp_prompt_global_task_noca}
This prompt allocates the daily time budget across pending tasks from all eligible repositories in one decision.

\begin{promptbox}
You are simulating this GitHub user in 2018.
Decide which pending tasks to complete today across ALL your repositories.

Profile:
\{background\}

Long-term memory:
\{long\_term\_memory\}

Recent tech news:
\{news\}

Today you have at most 480 minutes total to complete tasks across all repositories.

Below are all your repositories that currently have pending tasks:
\{repo\_blocks\}

Decide which tasks, if any, to complete today.
You may select tasks from one or multiple repositories.

Rules:
\begin{itemize}[label=-, leftmargin=1.5em, nosep]
\item Select zero or more tasks from the Pending Tasks listed above.
\item Copy each selected task's text exactly as shown.
\item For each selected task, write one concise and plausible Git commit message.
\item Estimate "time\_cost" as the actual completion time today, in integer minutes.
\item The total "time\_cost" across ALL selected tasks must not exceed 480 minutes.
\item Return an empty array if no task should be completed today.
\end{itemize}

Return ONLY a JSON array.
Each item in the array must contain exactly these fields:
\begin{itemize}[label=-, leftmargin=1.5em, nosep]
\item "repo\_id": integer repository ID
\item "task": one task string copied exactly from Pending Tasks
\item "commit\_message": one concise commit message for that task
\item "decision\_reason": a short explanation within 50 words
\item "time\_cost": the estimated completion time for that task in minutes as an integer
\end{itemize}
\end{promptbox}

\textbf{(12) OWNED\_REPO\_OTHERS\_COMMIT\_REVIEW\_\allowbreak{}PROMPT.}
\label{sec:supp_prompt_commit_review}
This prompt records whether CA is used while reviewing each listed collaborator commit.

\begin{promptbox}
You are simulating this GitHub user in 2018.
Decide whether to use CodeAgent when reviewing commits made by other contributors in your own repository.

Profile:
\{background\}

Long-term memory:
\{long\_term\_memory\}

Recent tech news:
\{news\}

Commits made by other collaborators that you will review today:
\{other\_collaborator\_commits\}

You are reviewing commits made by other collaborators in a repository that you own.

Rules:
\begin{itemize}[label=-, leftmargin=1.5em, nosep]
\item Every listed commit will be reviewed today.
\item Treat every listed commit as approved.
\item For each listed commit, decide only whether you would use CodeAgent during the review.
\item Make the CodeAgent decision based on your CodeAgent orientation, the commit itself, and the repository context.
\end{itemize}

Return ONLY a valid JSON array.
Each array element must be an object with exactly these fields:
\begin{itemize}[label=-, leftmargin=1.5em, nosep]
\item "commit\_id": the commit id corresponding to the review
\item "codeagent\_decision": either "YES" or "NO"
\item "decision\_reason": a short explanation, at most 50 words
\end{itemize}
\end{promptbox}
\textbf{(13) OWNED\_REPO\_OTHERS\_ISSUE\_RESOLUTION\_\allowbreak{}PROMPT\_WITH\_CA.}
\label{sec:supp_prompt_issue_resolution_ca}
This prompt assigns each collaborator issue a resolution state, produces the required commit or reply, and records whether CA is used. Conversation history supports continued resolution across days.

\begin{promptbox}
You are simulating this GitHub user in 2018.
Decide how to resolve issues opened by other collaborators in your own repository.

Profile:
\{background\}

Long-term memory:
\{long\_term\_memory\}

Recent tech news:
\{news\}

Issues opened by other collaborators that need your resolution decision today:
\{other\_collaborator\_issues\}

You are resolving issues opened by other collaborators in a repository that you own.

Some issues may already have a "Conversation history" from previous days — you and the issue opener may have exchanged messages before.
Read the full conversation history for each issue carefully before making your decision.

For each listed issue, decide:
1. Whether to use CodeAgent during the resolution.
2. The resolution outcome — one of three states:
\begin{itemize}[label=-, leftmargin=1.5em, nosep]
\item "fully\_resolved": You completely fixed the issue. Provide a commit message describing the fix.
\item "partially\_resolved": You made partial progress. Provide a commit message for what was done, AND an interaction message to the issue opener asking them to supplement information or clarify the remaining part.
\item "not\_solved": You could not solve it. Provide an interaction message to the issue opener asking for more information or explaining why it cannot be resolved yet.
\end{itemize}
3. An interaction message to the issue opener when needed:
\begin{itemize}[label=-, leftmargin=1.5em, nosep]
\item For "fully\_resolved": set interaction\_message to an empty string "" (no further communication needed).
\item For "partially\_resolved": write a message explaining what was done and what additional input you need from the opener.
\item For "not\_solved": write a message asking the opener for more details or explaining the blocker.
\end{itemize}

Important — conversation continuity:
\begin{itemize}[label=-, leftmargin=1.5em, nosep]
\item If the issue has conversation history and the opener provided new information, your resolution\_status CAN improve (e.g., not\_solved → partially\_resolved, or partially\_resolved → fully\_resolved).
\item If you are still waiting for information that the opener has not yet provided, you may repeat your request in the interaction\_message.
\item Each response should read as the next turn in an ongoing GitHub issue thread.
\end{itemize}

Rules:
\begin{itemize}[label=-, leftmargin=1.5em, nosep]
\item Make the CodeAgent decision based on your CodeAgent orientation, the issue itself, and the repository context.
\item commit\_message should be a concise Git commit message describing the resolution (for fully\_resolved and partially\_resolved; use empty string "" for not\_solved).
\item interaction\_message should be written as if you are replying to the issue opener on GitHub, in natural language. Use empty string "" when no interaction is needed.
\item decision\_reason should briefly explain your resolution choice and CodeAgent choice.
\end{itemize}

Return ONLY a JSON array.
Each array element must be an object with exactly these fields:
\begin{itemize}[label=-, leftmargin=1.5em, nosep]
\item "issue\_id": the issue id (matching the Issue ID in the issue listing above)
\item "codeagent\_decision": either "YES" or "NO"
\item "resolution\_status": one of "fully\_resolved", "partially\_resolved", "not\_solved"
\item "commit\_message": a concise Git commit message (empty string "" for not\_solved)
\item "interaction\_message": a reply message to the issue opener (empty string "" when fully\_resolved)
\item "decision\_reason": a short explanation, at most 50 words
\end{itemize}
\end{promptbox}
\textbf{(14) OWNED\_REPO\_OTHERS\_ISSUE\_RESOLUTION\_\allowbreak{}PROMPT\_WITHOUT\_CA.}
\label{sec:supp_prompt_issue_resolution_noca}
This variant uses the same resolution states and conversation history without returning a CA-decision field.

\begin{promptbox}
You are simulating this GitHub user in 2018.
Decide how to resolve issues opened by other collaborators in your own repository.

Profile:
\{background\}
\{tendency\_description\}

Long-term memory:
\{long\_term\_memory\}

Recent tech news:
\{news\}

Issues opened by other collaborators that need your resolution decision today:
\{other\_collaborator\_issues\}

You are resolving issues opened by other collaborators in a repository that you own.

Some issues may already have a "Conversation history" from previous days — you and the issue opener may have exchanged messages before.
Read the full conversation history for each issue carefully before making your decision.

For each listed issue, decide:
1. The resolution outcome — one of three states:
\begin{itemize}[label=-, leftmargin=1.5em, nosep]
\item "fully\_resolved": You completely fixed the issue. Provide a commit message describing the fix.
\item "partially\_resolved": You made partial progress. Provide a commit message for what was done, AND an interaction message to the issue opener asking them to supplement information or clarify the remaining part.
\item "not\_solved": You could not solve it. Provide an interaction message to the issue opener asking for more information or explaining why it cannot be resolved yet.
\end{itemize}
2. An interaction message to the issue opener when needed:
\begin{itemize}[label=-, leftmargin=1.5em, nosep]
\item For "fully\_resolved": set interaction\_message to an empty string "" (no further communication needed).
\item For "partially\_resolved": write a message explaining what was done and what additional input you need from the opener.
\item For "not\_solved": write a message asking the opener for more details or explaining the blocker.
\end{itemize}

Important — conversation continuity:
\begin{itemize}[label=-, leftmargin=1.5em, nosep]
\item If the issue has conversation history and the opener provided new information, your resolution\_status CAN improve (e.g., not\_solved → partially\_resolved, or partially\_resolved → fully\_resolved).
\item If you are still waiting for information that the opener has not yet provided, you may repeat your request in the interaction\_message.
\item Each response should read as the next turn in an ongoing GitHub issue thread.
\end{itemize}

Rules:
\begin{itemize}[label=-, leftmargin=1.5em, nosep]
\item commit\_message should be a concise Git commit message describing the resolution (for fully\_resolved and partially\_resolved; use empty string "" for not\_solved).
\item interaction\_message should be written as if you are replying to the issue opener on GitHub, in natural language. Use empty string "" when no interaction is needed.
\item decision\_reason should briefly explain your resolution choice.
\end{itemize}

Return ONLY a JSON array.
Each array element must be an object with exactly these fields:
\begin{itemize}[label=-, leftmargin=1.5em, nosep]
\item "issue\_id": the issue id (matching the Issue ID in the issue listing above)
\item "resolution\_status": one of "fully\_resolved", "partially\_resolved", "not\_solved"
\item "commit\_message": a concise Git commit message (empty string "" for not\_solved)
\item "interaction\_message": a reply message to the issue opener (empty string "" when fully\_resolved)
\item "decision\_reason": a short explanation, at most 50 words
\end{itemize}
\end{promptbox}
\textbf{(15) ENVIRONMENT\_MEMORY\_UPDATE\_PROMPT.}
\label{sec:supp_prompt_env_memory}
This prompt incorporates durable information from environment news into long-term memory while discarding transient details.

\begin{promptbox}
You are simulating this GitHub user in 2018.

Profile:
\{background\}

Existing long-term memory:
\{long\_term\_memory\}

News:
\{news\}

Update the qualitative long-term memory with information from the environment
that is likely to remain relevant to future GitHub behavior.

Keep:
\begin{itemize}[label=-, leftmargin=1.5em, nosep]
\item lasting changes in the technical or GitHub environment
\item external information that may continue to influence your attention, interests, or project choices
\item important new knowledge about tools, ecosystems, collaborators, or repository-related opportunities
\item durable changes in how CodeAgent may relate to your workflow, awareness, or future decisions
\end{itemize}

Do not keep:
\begin{itemize}[label=-, leftmargin=1.5em, nosep]
\item one-off details that are unlikely to matter later
\item raw news wording or verbatim environment logs
\item temporary noise
\item repeated information unless the news meaningfully updates it
\end{itemize}

Write concise factual statements.
Keep the memory consistent with the user's profile.
Maximum 200 words.

Return ONLY one valid JSON object with exactly these fields:
\begin{itemize}[label=-, leftmargin=1.5em, nosep]
\item "memory": the updated long-term memory text (max 200 words)
\item "reason": concise explanation within 50 words of why these specific updates were made, citing the news content that motivated each change
\end{itemize}

Do not output any extra text.
\end{promptbox}
\textbf{(16) MULTI\_BEHAVIOUR\_MEMORY\_PROMPT.}
\label{sec:supp_prompt_multi_behaviour}
This prompt updates long-term memory from the day's actions and repository context without retaining action-by-action logs.

\begin{promptbox}
You are simulating this GitHub user in 2018.

Profile:
\{background\}

Existing long-term memory:
\{long\_term\_memory\}

Today's context:
News:
\{news\}

Owned repositories:
\{owned\_repo\_info\}

Participated repositories:
\{participated\_repo\_info\}

Today's actions:
\{actions\}

Update the qualitative long-term memory with information likely to affect
future GitHub behavior.

Keep:
\begin{itemize}[label=-, leftmargin=1.5em, nosep]
\item stable projects and repository involvement
\item recurring programming-language or technical interests
\item lasting collaboration patterns and responsibilities
\item durable changes in CodeAgent awareness, adoption, or use
\end{itemize}

Do not keep:
\begin{itemize}[label=-, leftmargin=1.5em, nosep]
\item individual commits or commit messages
\item action-by-action logs
\item temporary news or one-off events
\item numerical activity statistics; these are maintained separately by Python
\item repeated information unless it has changed
\end{itemize}

Write concise factual statements.
Use only the supplied information.
Maximum 200 words.

Return ONLY one valid JSON object with exactly these fields:
\begin{itemize}[label=-, leftmargin=1.5em, nosep]
\item "memory": the updated long-term memory text (max 200 words)
\item "reason": concise explanation within 50 words of why these specific updates were made, citing evidence from today's context and actions
\end{itemize}

Do not output any extra text.
\end{promptbox}
\textbf{(17) TENDENCY\_UPDATED\_PROMPT\_WITH\_NEWS.}
\label{sec:supp_prompt_tendency_news}
This prompt updates the five-level CA tendency from external information and current memory.

\begin{promptbox}
You are simulating this GitHub user in 2018.
Update your CodeAgent orientation tendency based only on the information below.

Tendency scale (\{MIN\_RATING\}–\{MAX\_RATING\}):
\begin{itemize}[label=-, leftmargin=1.5em, nosep]
\item 1 = aware of CodeAgent, but rarely relies on it
\item 2 = occasionally relies on CodeAgent in limited situations
\item 3 = has a moderate tendency to rely on CodeAgent
\item 4 = often relies on CodeAgent
\item 5 = relies on CodeAgent very heavily
\end{itemize}

When updating the tendency:
\begin{itemize}[label=-, leftmargin=1.5em, nosep]
\item First determine whether the user is aware of CodeAgent at all.
\item Then determine how strongly CodeAgent appears to affect the user's workflow, attention, or GitHub behavior.
\item Use only evidence from the profile, long-term memory, latest news, and current tendency.
\item If there is no new evidence relevant to CodeAgent, keep the tendency unchanged.
\item Do not increase the tendency unless the text supports greater awareness, familiarity, use, or reliance.
\item Do not decrease the tendency unless the text supports weaker engagement, rejection, or loss of relevance.
\end{itemize}

Profile:
\{background\}

Current long-term memory:
\{long\_term\_memory\}

Latest news:
\{news\}

Current CodeAgent Tendency:
\{current\_tendency\}

Return ONLY one valid JSON object with exactly these fields:
\begin{itemize}[label=-, leftmargin=1.5em, nosep]
\item "tendency": an integer within the valid range
\item "reason": a concise explanation within 50 words describing why the tendency stayed the same or changed
\end{itemize}

Do not output any extra text.
\end{promptbox}
\textbf{(18) TENDENCY\_UPDATED\_PROMPT\_WITH\_\allowbreak{}RECORDS.}
\label{sec:supp_prompt_tendency_records}
This prompt updates the same tendency from the developer's direct behavioral records and CA-use decisions.

\begin{promptbox}
You are simulating this GitHub user in 2018.
Update your CodeAgent orientation tendency based only on the information below.

Tendency scale (\{MIN\_RATING\}–\{MAX\_RATING\}):
\begin{itemize}[label=-, leftmargin=1.5em, nosep]
\item 1 = aware of CodeAgent, but rarely relies on it
\item 2 = occasionally relies on CodeAgent in limited situations
\item 3 = has a moderate tendency to rely on CodeAgent
\item 4 = often relies on CodeAgent
\item 5 = relies on CodeAgent very heavily
\end{itemize}

When updating the tendency:
\begin{itemize}[label=-, leftmargin=1.5em, nosep]
\item First determine whether the user is aware of CodeAgent at all.
\item Then determine how strongly CodeAgent appears to affect the user's workflow, attention, or GitHub behavior.
\item Use only evidence from the profile, long-term memory, today's actions, latest news, and current tendency.
\item If there is no new evidence relevant to CodeAgent, keep the tendency unchanged.
\item Do not increase the tendency unless the text supports greater awareness, familiarity, use, or reliance.
\item Do not decrease the tendency unless the text supports weaker engagement, rejection, or loss of relevance.
\end{itemize}

Profile:
\{background\}

Current long-term memory:
\{long\_term\_memory\}

Recent actions you took today:
\{actions\}

Latest news:
\{news\}

Current CodeAgent Tendency:
\{current\_tendency\}

Return ONLY one valid JSON object with exactly these fields:
\begin{itemize}[label=-, leftmargin=1.5em, nosep]
\item "tendency": an integer within the valid range
\item "reason": a concise explanation within 50 words describing why the tendency stayed the same or changed
\end{itemize}

Do not output any extra text.
\end{promptbox}

\subsection{9. Metric Definitions and Statistical Procedures}
\label{sec:supp_metrics}

This section provides formal definitions, measurement units, and aggregation procedures for the metrics reported in the main Results section. Its purpose is to clarify what each metric measures and how it is calculated across four result parts, including simulation validity, production and adoption, developer interaction, and public knowledge. Unless otherwise stated, each metric is calculated separately for each simulation run and then aggregated across the three independent runs. This operation prevents differences in run-level activity volume from affecting the reported comparisons.

\subsubsection{9.1 Simulation Validity}
\label{sec:supp_validity_metrics}

\paragraph{(1) Validation Target and Unit of Analysis.}

Simulation validity is evaluated by comparing developer activities generated by the baseline simulation with activities observed during the Simulation period window. Real and simulated activities are aligned by developer, day, and behavior type. Simulated data are obtained from daily state snapshots and event logs, while real data are obtained from preprocessed GitHub activity records.

Let $\mathcal{M}$ denote the six developer behavior types, $N$ the number of aligned developers, and $D$ the number of days in the validation window. Let $r_{i,d,m}$ and $s_{i,d,m}$ denote the real and simulated counts, respectively, of behavior type $m$ produced by developer $i$ on day $d$. The daily total activity of each developer is defined as:

\begin{equation*}
y_{i,d}^{\mathrm{emp}}=\sum_{m\in\mathcal{M}}r_{i,d,m},
\qquad
y_{i,d}^{\mathrm{sim}}=\sum_{m\in\mathcal{M}}s_{i,d,m}.
\end{equation*}

\paragraph{(2) Activity-Volume Errors.}

The MAE and RMSE reported in the main text are calculated at the agent--day level. They assess whether the simulation reproduces both the overall level and daily pattern of developer activity:

\begin{equation*}
\mathrm{MAE}=\frac{1}{ND}\sum_{i=1}^{N}\sum_{d=1}^{D}\left|y_{i,d}^{\mathrm{sim}}-y_{i,d}^{\mathrm{emp}}\right|,
\end{equation*}

\begin{equation*}
\mathrm{RMSE}=\sqrt{\frac{1}{ND}\sum_{i=1}^{N}\sum_{d=1}^{D}\left(y_{i,d}^{\mathrm{sim}}-y_{i,d}^{\mathrm{emp}}\right)^2}.
\end{equation*}

MAE measures the average absolute deviation, whereas RMSE assigns greater weight to large daily errors. The mean and median total activities are calculated from each developer's cumulative activity, $\sum_d y_{i,d}$, over the complete validation window and are used to compare overall production levels.

\paragraph{(3) Developer Heterogeneity and $\Delta$Gini.}

Beyond average activity levels, we examine whether the simulation preserves differences across developers. For a nonnegative sequence ordered as $x_1\leq\cdots\leq x_n$, the Gini coefficient is defined as:

\begin{equation*}
G(x)=\frac{2\sum_{i=1}^{n}i x_i}{n\sum_{i=1}^{n}x_i}-\frac{n+1}{n}.
\end{equation*}

We set $G(x)=0$ when $n=0$ or $\sum_i x_i=0$. Gini coefficients are calculated for the real and simulated data along three dimensions:

\begin{itemize}
    \item \textbf{Activity Gini:} Let $c_i$ denote the total number of activities produced by developer $i$ during the validation window. Then, $G_{\mathrm{activity}}=G(\{c_i\}_{i=1}^{N})$.
    \item \textbf{Active-Repository Gini:} Let $r_i=|\{k:c_{i,k}\geq1\}|$ denote the number of repositories in which developer $i$ produces at least one activity. Then, $G_{\mathrm{active\_repo}}=G(\{r_i\}_{i=1}^{N})$.
    \item \textbf{Repository-Concentration Gini:} We first calculate the concentration of each developer's activities across repositories as $g_i=G(\{c_{i,k}:k\in\mathcal{R}_i\})$. We set $g_i=0$ when the developer is active in at most one repository. We then calculate $G_{\mathrm{conc}}=G(\{g_i\}_{i=1}^{N})$ to measure differences in repository-concentration patterns across developers.
\end{itemize}

For each dimension $X\in\{\mathrm{activity},\mathrm{active\_repo},\mathrm{conc}\}$, the difference between the real and simulated distributions is:

\begin{equation*}
\Delta_X=\left|G_X^{\mathrm{emp}}-G_X^{\mathrm{sim}}\right|.
\end{equation*}

The overall distributional deviation is:

\begin{equation*}
\Delta\mathrm{Gini}=\frac{1}{3}\left(\Delta_{\mathrm{activity}}+\Delta_{\mathrm{active\_repo}}+\Delta_{\mathrm{conc}}\right).
\end{equation*}

A smaller $\Delta\mathrm{Gini}$ indicates that the simulation better preserves developer heterogeneity in the real community.

\subsubsection{9.2 RQ1: Production, Efficiency, and Adoption}
\label{sec:supp_rq1_metrics}

\paragraph{(1) Planned Tasks and Completed Tasks.}

\textbf{Planned Tasks} are work items generated during daily task planning and added to a repository's pending-task list. Each task contains a description, creation date, and target repository. \textbf{Completed Tasks} are tasks subsequently selected and completed by an agent. Each completed task produces a corresponding commit record and is removed from the pending-task list.

The number of planned tasks on day $d$ equals the number of new task-creation events on that day, while the number of completed tasks equals the number of task-completion events. Cumulative values are obtained by summing daily events. The calculation uses task events rather than repeated state snapshots to avoid counting the same task more than once.

\paragraph{(2) Task Completion Time.}

Each completed task has a \texttt{time\_cost}, measured in simulated minutes, representing the agent's estimate of the time required to complete the task. For a CA-assisted task, this value represents the estimated completion time after CA support rather than the unassisted time. The median and quartiles reported in the main text are calculated from the \texttt{time\_cost} distribution of all completed tasks.

\paragraph{(3) CA Adoption.}

A developer is considered an adopter after actively choosing to use the Coding Agent in at least one task, commit review, or issue resolution activity. Let $U_d$ denote the set of developers who have used the CA at least once by day $d$. The cumulative adoption rate is defined as:

\begin{equation*}
\mathrm{Adoption}_d=\frac{|U_d|}{N}.
\end{equation*}

For completeness, we also define CA awareness as the share of developers who have learned about the Coding Agent, regardless of whether they have used it. Let $A_d$ denote the set of developers whose state is either \texttt{"Know CodeAgent"} or \texttt{"Use CodeAgent"} by day $d$. The cumulative awareness rate is:

\begin{equation*}
\mathrm{Awareness}_d=\frac{|A_d|}{N}.
\end{equation*}

Awareness measures the reach of CA-related information, whereas adoption measures actual tool use. The difference between the two captures the transition from learning about the tool to using it in practice. CA Awareness is included here as a supplementary measure and is not reported as part of the core metric set in the main text.

\paragraph{(4) CA-Assisted Commit Ratio.}

The daily CA-assisted commit ratio is the proportion of commits completed with CA support:

\begin{equation*}
\mathrm{CA\ Ratio}_d=
\frac{|\{q:q\text{ occurs on day }d,\ a(q)=1\}|}
{|\{q:q\text{ occurs on day }d\}|}.
\end{equation*}

The ratio is not calculated on days with no commits. It measures the penetration of CA support into community production and indicates whether a small group of adopters produces a disproportionate share of CA-assisted output.

\subsubsection{9.3 RQ2: Task Modes and Developer Interaction}
\label{sec:supp_rq2_metrics}

\paragraph{(1) Task-Mode Classification.}

Each completed commit task is denoted by $q$ and described by three variables:

\begin{itemize}
    \item $c(q)$: the developer who submits the task.
    \item $r(q)$: the owner of the target repository.
    \item $a(q)$: the CA-participation indicator, where $a(q)=1$ if the task uses the coding agent and $a(q)=0$ otherwise.
\end{itemize}

Tasks are divided into four modes according to whether they cross developer boundaries and whether the coding agent is involved:

\begin{table*}[htbp]
    \centering
    \small
    \caption{Formal definitions of the four task-execution modes.}
    \label{tab:interaction_modes}
    \begin{tabular}{l|l|l}
        \hline
        Mode & Condition & Interpretation \\
        \hline
        HHI & $c(q)\neq r(q)\land a(q)=0$ & Direct cross-developer task without CA support \\
        HSA & $c(q)=r(q)\land a(q)=0$ & Developer self-loop task without CA support \\
        AHI & $c(q)\neq r(q)\land a(q)=1$ & CA-assisted cross-developer task \\
        ASA & $c(q)=r(q)\land a(q)=1$ & CA-assisted developer self-loop task \\
        \hline
    \end{tabular}
\end{table*}

The four modes form a complete $2\times2$ classification based on cross-developer involvement and CA participation.

\paragraph{(2) Task Composition and Agent Mediation Rate.}

Let $Q_t$ denote all completed tasks during period $t$, and let $Q_t^m$ denote the subset belonging to mode $m$. The share of mode $m$ is:

\begin{equation*}
\mathrm{Share}_t^m=\frac{|Q_t^m|}{|Q_t|},
\qquad
m\in\{\mathrm{HHI},\mathrm{HSA},\mathrm{AHI},\mathrm{ASA}\}.
\end{equation*}

The proportion of cross-developer tasks involving CA assistance is defined as:

\begin{equation*}
\mathrm{AMR}_t=
\frac{|Q_t^{\mathrm{AHI}}|}
{|Q_t^{\mathrm{HHI}}|+|Q_t^{\mathrm{AHI}}|}.
\end{equation*}

This metric indicates how often CA support is involved in cross-developer tasks. 

\paragraph{(3) Context Breadth.}

Let $D_t$ denote the set of developer pairs involved in at least one cross-developer task during period $t$, and let $C_{ij,t}$ denote the distinct $(\mathrm{repository},\mathrm{mode})$ combinations involving pair $(i,j)$. Context Breadth is defined as:

\begin{equation*}
\mathrm{Context\ Breadth}_t=
\frac{1}{|D_t|}
\sum_{(i,j)\in D_t}|C_{ij,t}|.
\end{equation*}

A higher value indicates that a developer pair collaborates across more repositories or task modes, reflecting greater contextual breadth in the relationship.

\paragraph{(4) Repeated Interaction Rate.}

Let $n_{ij,t}^{\mathrm{task}}$ denote the number of cross-developer tasks involving pair $(i,j)$ during period $t$, and let $n_{ij,t}^{\mathrm{day}}$ denote the number of distinct days on which these tasks occur. The Repeated Interaction Rate is:

\begin{equation*}
\mathrm{Repeated\ Rate}_t=
\frac{
\sum_{(i,j)\in D_t}
\mathbb{I}\left(
n_{ij,t}^{\mathrm{task}}\geq2
\land
n_{ij,t}^{\mathrm{day}}\geq2
\right)}
{|D_t|}.
\end{equation*}

A developer pair is considered recurrent only when it interacts through at least two tasks on at least two different days. This metric captures short-term continuity in collaboration, while Context Breadth captures the range of contexts in which the relationship occurs.

\subsubsection{9.4 RQ3: Public Knowledge}
\label{sec:supp_rq3_metrics}

\paragraph{(1) Public Knowledge Corpus.}

The public knowledge corpus $\mathcal{C}$ contains simulation records that can be observed and reused by other developers:

\begin{enumerate}
    \item \textbf{Commit Messages:} Public descriptions of code changes.
    \item \textbf{Task Descriptions:} Task descriptions recorded in platform or repository states.
    \item \textbf{Issue Records:} Issue content, resolution states, and developer replies.
    \item \textbf{Repository Actions:} Descriptive records associated with repository creation and forking.
\end{enumerate}

The corpus is truncated at the end of the Warmup Period and the Simulation Period to compare the public knowledge accumulated at each stage. Non-English records are translated into English before preprocessing. All texts are converted to lowercase, with URLs, nonalphabetic characters, and repeated spaces removed. The processed records are then represented using TF-IDF vectors, retaining all terms that appear at least once.

\paragraph{(2) Knowledge Queries and Corpus-Size Control.}

The knowledge-query set $\mathcal{Q}$ is constructed from real GitHub activities occurring after the Simulation period and represents technical information that developers may subsequently need to retrieve. The source data contain 8,822 commits from March 19 to May 19, 2018, which are aggregated into 808 repository-level query texts.

The CA corpus generally contains more records than the real corpus. To prevent corpus size alone from increasing retrieval performance, we randomly sample the CA corpus to match the number of records in the real corpus. This procedure is repeated five times, and the mean and standard deviation are reported.

\paragraph{(3) Public Knowledge Coverage.}

Let $\operatorname{sim}(q,c)$ denote the TF-IDF cosine similarity between query $q$ and corpus record $c$. Given the threshold $\tau=0.3$, Public Knowledge Coverage is defined as:

\begin{equation*}
\mathrm{PKC}(\mathcal{Q},\mathcal{C},\tau)=
\frac{
|\{q\in\mathcal{Q}:\max_{c\in\mathcal{C}}\operatorname{sim}(q,c)>\tau\}|
}
{|\mathcal{Q}|}.
\end{equation*}

PKC is the proportion of queries for which at least one related trace can be found in the existing public records.


\paragraph{(4) Multi-Step Retrieval Process.}

To model how a newcomer searches public records without prior knowledge of the project structure, we use an iterative retrieval procedure with at most $S=10$ steps. For each query $q$:

\begin{enumerate}
    \item The current query retrieves the five most similar records that have not been viewed previously.
    \item If the highest cosine similarity exceeds $\tau=0.3$, retrieval is considered successful, and the current step is recorded.
    \item Otherwise, the eight highest-weighted new TF-IDF terms are extracted from the best-matching record and added to the query.
    \item Retrieval continues with the expanded query until a match is found or the ten-step limit is reached.
\end{enumerate}

This process treats public records not only as possible answers but also as intermediate cues that help newcomers formulate later searches.

\paragraph{(5) Retrieval Success Rate and Average Retrieval Steps.}

Let $z_q=1$ if query $q$ retrieves a record above the similarity threshold within $S$ steps and $z_q=0$ otherwise. Retrieval Success Rate is defined as:

\begin{equation*}
\mathrm{RSR}=
\frac{1}{|\mathcal{Q}|}
\sum_{q\in\mathcal{Q}}z_q.
\end{equation*}

Let $s_q$ denote the number of steps required for a successful query. For unsuccessful queries, we set $s_q=S$. Average Retrieval Steps is:

\begin{equation*}
\mathrm{ARS}=
\frac{1}{|\mathcal{Q}|}
\sum_{q\in\mathcal{Q}}s_q.
\end{equation*}

The main analysis uses a fixed random sample of 300 repository-level queries and applies the same queries across conditions. RSR measures whether relevant knowledge can be found, whereas ARS measures the effort required to locate it.

\end{document}